\documentclass[preprint,12pt]{elsarticle}

\usepackage{amssymb}
\usepackage{xcolor}
\usepackage{float}
\usepackage{amsmath}
\usepackage{graphicx}
\usepackage{subcaption}
\usepackage{caption}
\usepackage{subcaption}
\usepackage{lineno}

\usepackage{lineno}
\usepackage[hyperfootnotes=false]{hyperref}
\hypersetup{
	colorlinks,
	citecolor=false,
	linkcolor=Red,
	urlcolor=Blue}

\journal{Journal Name}

\begin{document}

\begin{frontmatter}



\title{COVID-19 personal protective equipment detection using real-time deep learning methods}


\author[1]{Shayan Khosravipour}
\author[1]{Erfan Taghvaei}
\author[1]{Nasrollah Moghadam Charkari}

\address[1]{Faculty of Electrical and Computer Engineering}
\address[1]{Tarbiat Modares University}
\address[1]{Tehran. Iran}

\begin{abstract}

The exponential spread of COVID-19 in over 215 countries has led WHO to recommend face masks and gloves for a safe return to school or work. We used artificial intelligence and deep learning algorithms for automatic face masks and gloves detection in public areas. We investigated and assessed the efficacy of two popular deep learning algorithms of YOLO (You Only Look Once) and SSD MobileNet for the detection and proper wearing of face masks and gloves trained over a data set of 8250 images imported from the internet. YOLOv3 is implemented using the DarkNet framework, and the SSD MobileNet algorithm is applied for the development of accurate object detection. The proposed models have been developed to provide accurate multi-class detection (Mask vs. No-Mask vs. Gloves vs. No-Gloves vs. Improper). When people wear their masks improperly, the method detects them as an improper class. The introduced models provide accuracies of (90.6\% for YOLO and 85.5\% for SSD) for multi-class detection. The systems' results indicate the efficiency and validity of detecting people who do not wear masks and gloves in public.
\end{abstract}


\end{frontmatter}


\section{Introduction}
\label{S:1}
Today, AI has a crucial role in every aspect of the COVID-19 crisis response. AI is composed of different techniques that are used as non-clinical approaches to mitigate the huge burden of health care systems. One of its roles is to prepare an assisting tool to prevent the spread of the virus through automatic tracing and surveillance of people who do not wear masks and gloves in the public area. When the virus was at its early stages, researchers quickly provided the necessary information and data for academic purposes. Among them were AI researchers, making their efforts to develop smart applications to overcome the limitations that humans pose against COVID-19 \cite{fiesco2020contributions}. A notable example is a deep learning system that uses the DarkNet model classifier. It can classify COVID-19 cases from raw chest X-ray images with an accuracy of 98.08 \cite{ozturk2020automated}. Since prevention is better than treatment, the COVID-19 mask and gloves detection system is a useful and feasible solution to mitigate this virus's spread. Almost the majority of countries in the world are going through a pandemic. At the time of writing this paper, more than 23 million COVID-19 cases have been confirmed in more than 215 countries, and the virus has caused more than 800 thousand
deaths (\url{https://www.worldometers.info/coronavirus}), the virus's growth rate is exponential in more populated countries which is discussed by regression coefficient of the log growth time-series data (i.e., number of new people infected per day) \cite{coelho2020global}. It is mainly
transmitted through human-to-human interaction \cite{Hu_2020}. It can spread through large droplets existing in the air when someone who carries the virus coughs, sneezes, or touches the face with a hand exposed by the virus. The respiratory tract is the gateway for the virus to enter the human body \cite{del2020covid},\cite{xiao2020evidence}. In addition to patients with visible symptoms, there are numerous
asymptomatic carrier cases with normal chest CT images and no self-reported fever, but they turned
out to be carriers of the virus \cite{bai2020presumed}. Unfortunately, due to the imbalance between supply and demand, some authorities attempt to lessen the importance of using masks \cite{li2020facial}. While there are no randomized controlled trials (RCT) for using masks as source control for SARS-CoV-2, numerous studies indicate that using personal protective
equipment (PPE), along with social distancing and personal hygiene, are
necessary to prevent the virus from entering the body through
infectious respiratory droplets and help flatten-the-curve. Furthermore, fitting gloves can prevent microorganisms obtained on the hands during daily tasks
and when known and unknown contaminated equipment or surfaces come into contact. As a result, the transmission rate of the virus is less significant, and the case-fatality rates (CFR) decreases \cite{howard2020face},\cite{centers2020interim},\cite{world2004practical}. One of the main
reasons that morbidity and mortality are higher in men than women is women tend to use
facial protection more than men \cite{bwire2020coronavirus}. Since the virus remains silent in the body, and the symptoms can go unnoticed for weeks, wearing protection is crucial to stop "Silent Spreaders," transmitting the virus \cite{zhang2020asymptomatic}. In a study undertaken in Hong Kong, findings from the Bonferroni-Dunn test with an adjusted level of significance indicate that older participants used a face mask less often than young people when they had respiratory symptoms. Also, the intention of citizens in wearing a mask was more about protecting others than themselves \cite{lee2020practice}. Even if there are carriers of the virus in
public areas, the virus's transmission rate becomes very low if they wear
a face mask. To further emphasize the importance of wearing facial protections and
providing a barrier between the face and the virus, the commitment of citizens to wearing
masks in public places contributed a great role in decreasing the spread
of the contagious virus in Vietnam,  in contrast to Brazil, were lacking sufficient protection amongst travelers in the airports, and not paying adequate attention to the WHO guidelines, were the main reasons the virus entered and spread throughout the whole country \cite{Ribeiro_2020}. All these statistics indicate that a positive outcome could be achieved when a high percentage of society collaborates and follows the safety guidelines \cite{thoi2020ho}. 

In addition to masks, in research that compared surface stability of SARS-CoV-2 with SARS-CoV-1, the virus potential to exist on surfaces up to days depending on the surface material has been indicated \cite{van2020aerosol}. The human hands come in contact with these unclean surfaces every day. When the individual does not wear a glove, hands can be contaminated by the virus and results in infection upon touching the facial organs. In this regard, Wearing masks and gloves are essential for the safety of healthcare workers (HCW) and other staff that are at high risk of getting infected and spreading the virus \cite{heinzerling2020transmission},\cite{soetikno2020considerations}. 

To conclude, numerous studies suggest practical solutions to help reduce the spread of the virus, such as alerting the public and urging them to use personal protective equipment, which results in spread prevention, as well as a decrease in public anxiety \cite{li2020substantial}\cite{chang2020covid}. Thus, people who are willing to visit public places during this severe pandemic should follow the safety guidelines if the communities aim to be victorious in the battle against COVID-19.

In this research, a method to track and supervise the proper enforcement of health recommendations for preventing the COVID-19 pandemic based on deep learning is proposed. Deep learning is one of the branches of AI that works like the human brain with many neurons. The word deep derives from the expansion of the network size, which is proportionate to the number of layers \cite{lecun2015deep} \cite{krizhevsky2012imagenet}. Convolutional neural networks (CNN) produce a state of the art results on image and video data.  CNN consists of a series of convolutional layers. In convolutional layers, multiple kernels convolve with input to produce a feature map; this layer function can be expressed using (1).  K\textsubscript{ij} denotes the convolutional kernel,  L\textsubscript{input} and L\textsubscript{output} represent the number of input and output features, respectively.

\[ \scalebox{1.2}{1) feature-map\textsubscript{i} = $\sum_{j=1}^{L\textsubscript{input}} K\textsubscript{ij} \ast input\textsubscript{j} + bias\textsubscript{j}\quad 1\leqslant{i}\leqslant{L\textsubscript{output}}$} \]

Then, this feature map passes through the activation function to introduce non-linearity into the output. $\Psi$ could be any nonlinear function like tanh, relu, etc. in formula (2).

\[ \scalebox{1.2}{2) $output\textsubscript{i} = \Psi$(feature-map\textsubscript{i})} \]

\begin{figure}[H]
	\centering\includegraphics[width=0.7\linewidth]{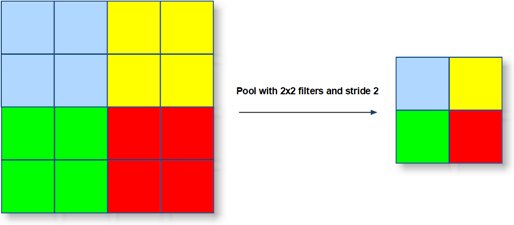}
	\caption{Pooling function}
	\label{fig:pooling}
\end{figure}

Finally, an m $\times$ m filter with a stride n is applied to the input vector and outputs maximum or minimum or average values of each subarea called pooling. It decreases the input's spatial size to reduce the number of parameters and computation in the network. The functionality of pooling layer is shown in figure \ref{fig:pooling}.

The differences between CNNs come from the number of convolutional layers, pooling function, and other internal parameters to accomplish a specific task like object detection or classification. In this regard, FRCNN \cite{ren2015faster}, Mask-RCNN \cite{he2017mask},YOLO \cite{redmon2016you},SSD \cite{liu2016ssd}
are the most well-known detection algorithms in the object detection area. Since real-time detection is crucial in this work, YOLOv3 and SSD MobileNet are selected. After training the custom dataset on these two deep learning models, their effectiveness has been compared. These models have a similar end-to-end architecture, computing a feature map with running a convolutional network on input image only once, which results in detecting objects in real-time with good accuracy. The YOLOv3 applies a new network, Darknet-53, whereas our SSD uses the MobileNet network \cite{howard2017mobilenets}. In societies where there is not an abundant supply of GPU power, SSD is recommended.

The experiments' results indicate the effectiveness and validity of the mentioned methods in supervising people to observe the regulation of using protectives in public areas.  
\section{YOLO: You only look once}
YOLO is one of the best object detection models which detect objects in real-time and provides a good trade-off between speed and accuracy. YOLO can retrieve contextual information about the classes as it observes the entire image during training and test. In contrast, the R-CNN models need a separate stage to fetch the target region \cite{buric2018ball}. Finally, it is generalizable since it can detect one object in various poses. There are three official versions of YOLO, where we have employed the latest version in the paper.

\begin{figure}[H]
	\centering\includegraphics[width=1\linewidth]{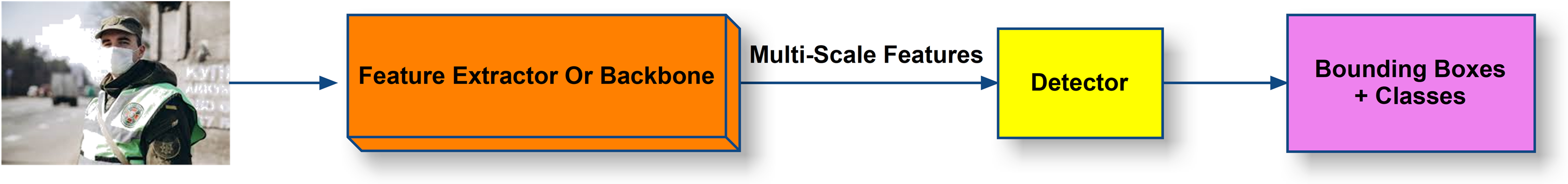}
	\caption{A high-level representation of the object detection methods}
	\label{fig:fianeltoloeeef}
\end{figure}

A high-level diagram of an object detector is shown in figure \ref{fig:fianeltoloeeef}. The whole system is composed of two major components: Feature Extractor and Detector. When an image comes in, it passes through the feature extractor first, and feature embedding is obtained at different scales. Then, these features are feed into branches of the detector to get bounding boxes and class information.

\begin{figure}[H]
	\centering\includegraphics[width=1.\linewidth]{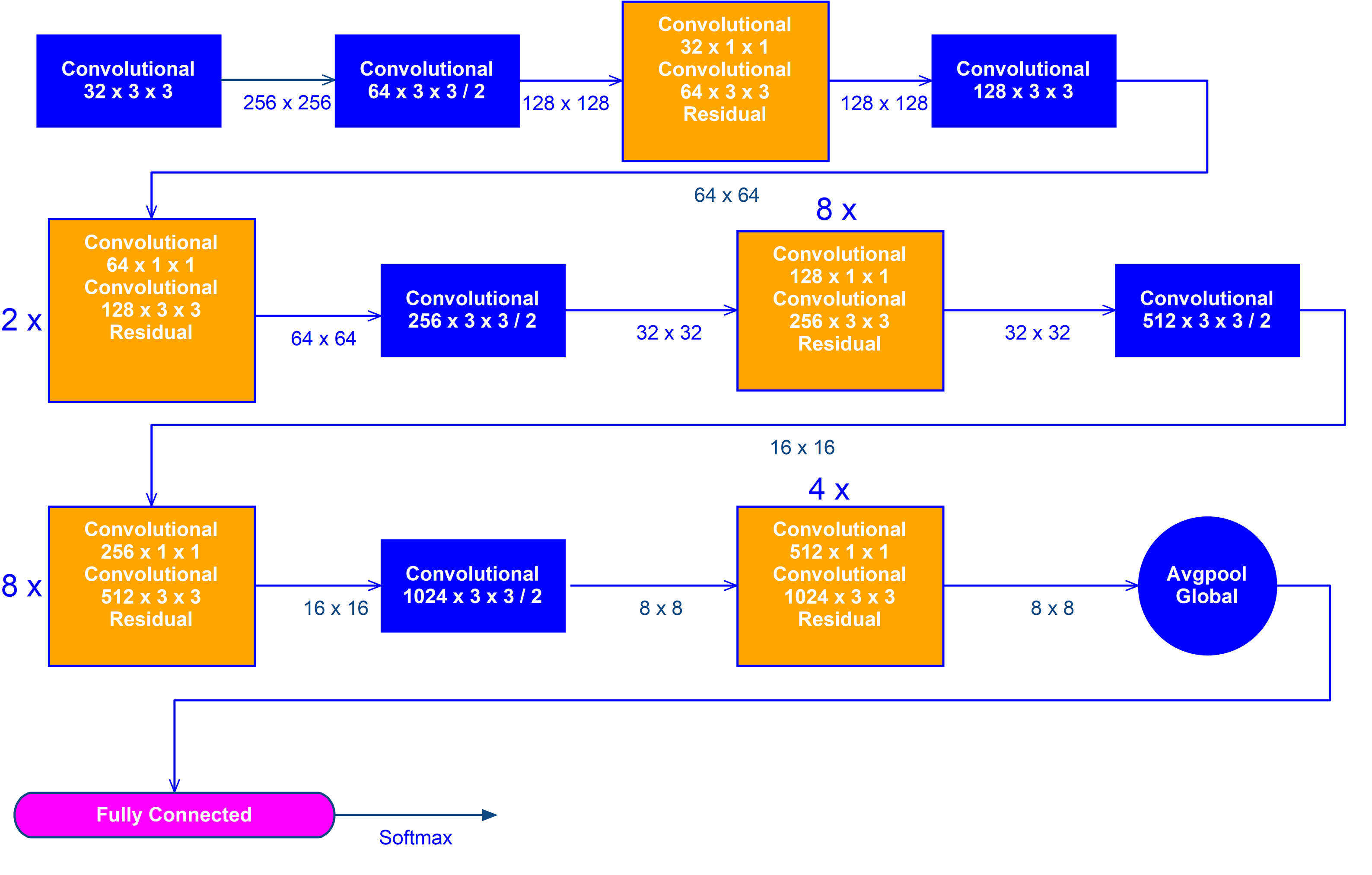}
	\caption{Darknet53 feature extractor architecture}
	\label{fig:darknet}
\end{figure}

\begin{figure}[H]
	\centering\includegraphics[width=1.1\linewidth]{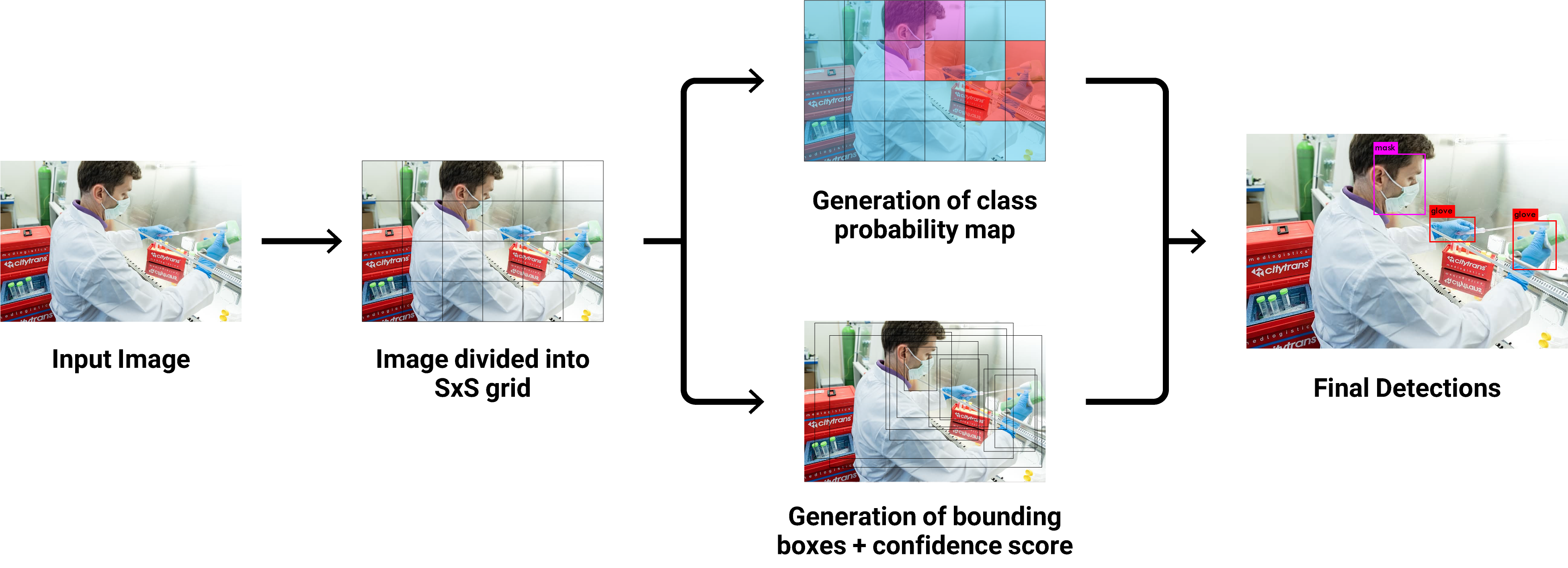}
	\caption{YOLO object detection process}
	\label{fig:grid1}
\end{figure}

YOLOv3 \cite{redmon2018yolov3} uses a new network Darknet-53 as a feature extractor which is shown in figure \ref{fig:darknet}, Darknet-53 has 53 convolutional layers. It is much deeper than the old version, which has only 19 convolutional layers \cite{redmon2017yolo9000}, performing detections at three different scales, improving the accuracy of almost 9.8\% in $608\times608$ image size. However, YOLOv2 performs at a better speed due to a lighter architecture. Moreover, it has residuals or shortcut connections to allow the gradients to flow through the network without passing through the activation functions. By adding Feature Pyramids in YOLOv3, the models' accuracy has improved in detecting small objects.
The stages of the YOLO detection model are shown in figure \ref{fig:grid1}. In the first step,
the model divides the image into an S $\times$ S grid. Then, the grid containing the center of the ground truth bounding box of an object is activated for detecting the object. Each grid is responsible for predicting B bounding boxes, their confidence scores, and C conditional probabilities for classes \cite{redmon2018yolov3}.

\begin{figure}[H]
	\centering\includegraphics[width=0.6\linewidth]{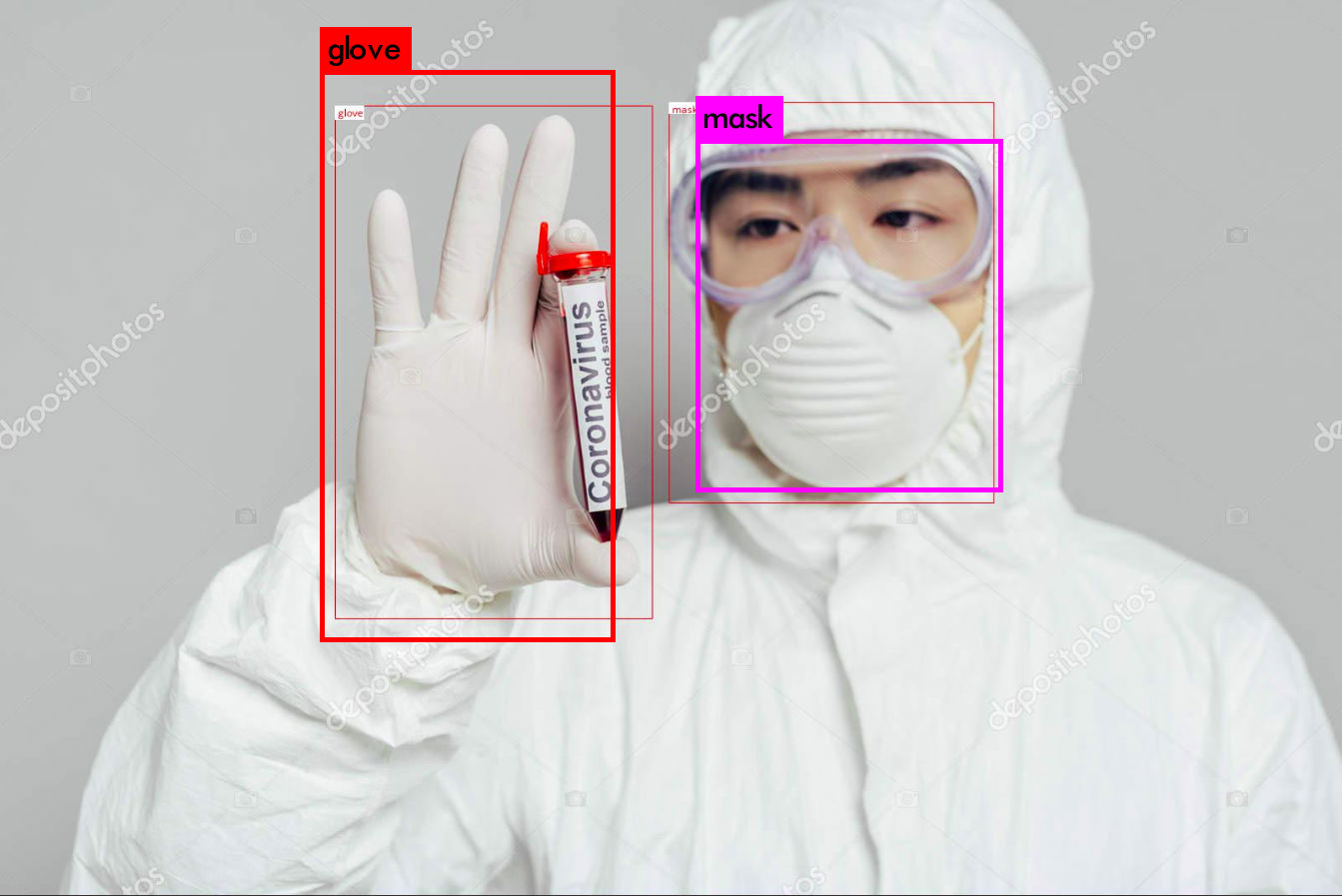}
	\caption{YOLOv3 aligns very good with the target object. Thick bounding boxes belong to YOLOv3
		and the thin boxes with small name sizes are the ground truth bounding boxes.}
	\label{fig:ioucompare}
\end{figure}

Same as all other detectors, the performance of YOLOv3 decreases as the IOU threshold increases. The Mean average precision on more than 50\% of Intersection over Union is a judgment metric used to check an object detector's accuracy on a particular dataset. The ground-truth bounding boxes (hand-labeled bounding boxes) and the predicted bounding boxes from the detector model are the two main factors used to find an object detector's accuracy. The IOU metric is computed by the ratio between the overlapped areas and the areas of Union. Refer to figure \ref{fig:ioucompare}, the thin rectangles are the ground truth boxes, and the thick boxes are the objects detected by YOLOv3. As observed, an appropriate result is provided. Consequently, there is no doubt that YOLOv3 is one of the best state-of-the-art object detectors available, having an acceptable trade-off between accuracy and speed. In this paper YOLOv3 has been used to solve the five-class detection problem, the detection kernels shape is (1 $\times$ 1 $\times$ 30) where 30 comes from 30 = ((k\textsubscript{1}+k\textsubscript{2}+k\textsubscript{3})$\times$3).\\
k\textsubscript{1} = 5\qquad     The number of classes (Mask,Improper,No-mask,Glove,No-glove)\\
k\textsubscript{2} = 4\qquad     The bounding boxes attribute(x-center,y-center,width,height)\\
k\textsubscript{3} = 1\qquad     The presence of object \\
Finally, 3 is the default value for the number of bounding boxes a cell on the feature map can predict.\\
\section{SSD: Single Shot MultiBox Detector}
Similar to YOLO, SSD detects objects in a single deep learning network. After producing prediction rates to determine the likelihood of an object's existence in a bounding box, SSD makes the necessary adjustments to better shape the targeted object in the bounding box. Since SSD uses the predictions from various activation maps, it handles images with various sizes properly. Both YOLO and SSD use a final non-maximum-suppression step in the final detection stage. One of the main characteristics of SSD is its capability in detecting larger objects. Furthermore, the performance of SSD is less sensitive to the quality of the feature extractor than Faster R-CNN and R-FCN. However, it does not perform very well on small objects \cite{huang2017speed}. 

\begin{figure}[H]
	\centering\includegraphics[width=1.\linewidth]{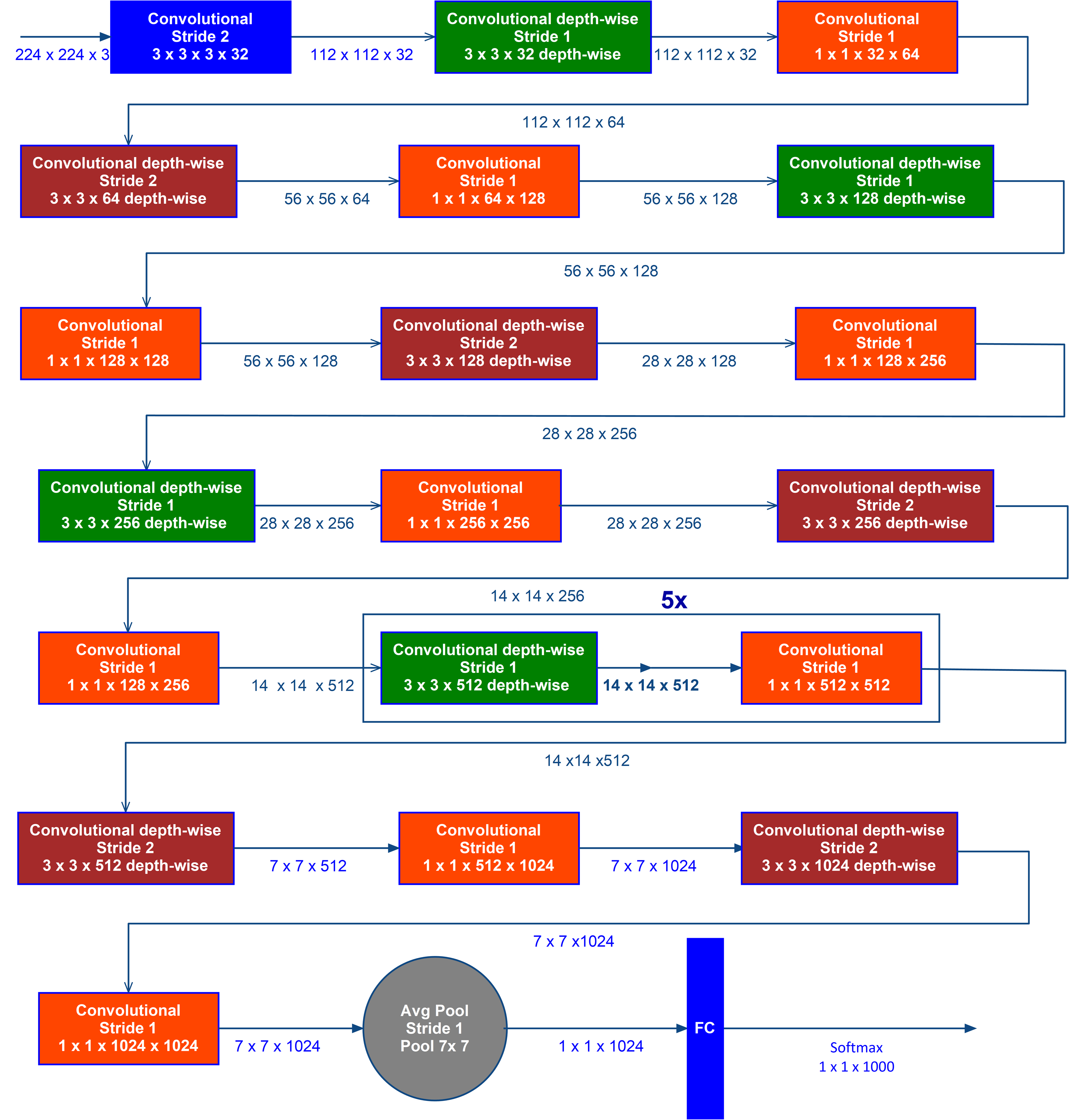}
	\caption{Our SSD feature extractor (MobileNet architecture)}
	\label{fig:mobilenet}
\end{figure}

Comparing SSD and YOLOv3 performance, with a fixed size of the objects, YOLOv3 outperformed SSD in both accuracy and speed \cite{park2019automated}. In this study, the dataset comprised of all kinds of object size. Thus, the detector should identify small, large, near, and far objects in the image. The MobileNet architecture's objective, which is shown in figure \ref{fig:mobilenet}, is to make neural networks lighter and portable for mobile and embedded applications, which is maintained primarily by depth-wise separable convolutions \cite{howard2017mobilenets}.
Like other object detectors, SSD has two major parts: backbone and head. The backbone part is a pre-trained image classification network for feature extraction. Here, we have used MobileNet architecture as a feature extractor because of its speed and proper accuracy, which is the result of using a combination of normal convolution and depthwise convolution. DW convolution is performed independently for each of the input channels. It significantly reduces the computational cost by omitting convolution in the channel domain. The SSD consists of one or more convolutional layers added to the MobileNet. The outputs are translated as the bounding boxes and classes of objects within the spatial area of the final layers activation.

The main difference between YOLO and SSD is that SSD uses multi-scale convolutional feature maps at the top of the network, but YOLO uses fully connected layers.

Some extra augmentation e.g., a technique that can be used for the artificial change of the size and other characteristics of a training image, and other techniques such as: converting RGB to gray, vertical flip, 90-degree rotation, and adding Gaussian noise, have been applied to get closer to YOLO results.
\section{Experiments}
In this section, to evaluate our proposed methods' robustness, our test datasets have been used to compare SSD and YOLOv3 results.
\subsection{Experiment Setup}
The dataset has been split randomly into a train, validation, and test set with 4250, 2000, and 2000 images individually in the experiments. The other learning parameters for both methods are listed in table \ref{tab:parameters}.

\begin{table}[H]
	\centering
	\begin{tabular}{l l l}
		\hline
		\textbf{training parameter} & \textbf{YOLOv3} & \textbf{SSD}\\
		\hline
		optimizer & Stochastic gradient descent & Stochastic gradient descent \\
		learning-rate & 0.001 & 0.0002 \\
		momentum & 0.9 & 0.9  \\
		epoch & 76 & 155\\
		training-set & 4250 & 4250\\
		validation-set & 2000 & 2000\\
        test-set & 2000 & 2000\\
		batch-size & 64 & 16\\
		image-size & 416$\times$416 & 300$\times$300\\
	    \hline
	\end{tabular}
	\caption{learning parameters}
	\label{tab:parameters}
\end{table}

\subsection{COVID-19 protective Dataset}
A set of human images, with and without protective equipment, as explained in the following. A total number of 8250 image data was carefully selected and collected for our purpose. Google Images were the source for gathering the required data. Masks with various forms, colors, and styles exist in the dataset. Due to the increased popularity of face shields, pictures with people wearing face shields with the intent to be detected as masks were also included. The diverse dataset comprises of images consisting of only a few numbers of people and images from crowded places, each with different backgrounds. As a result, objects with different sizes and qualities have been labeled in the annotation process. After the hand labeling process was complete, the number of image classes are brought in table \ref{tab:result2}, which are 34942 objects in total. Due to the reason that the number of gloves and no-gloves were incommensurate, up-sampling was done by repetition to balance out the dataset. For the test phase, after training the data and producing the necessary weights, 2000 images were randomly selected from the dataset to test the methods' accuracy. The dataset could be divided into four main groups: Ideal, masks but no gloves, improper mask-wearing, and poor hygiene. The ideal group is the people wearing both masks and gloves and, as a result, having perfect hygiene. The second group is protecting their facial organs by wearing masks; however, they do not wear gloves. The third group is those wearing the masks improperly, and finally, the last group is the people not wearing any protection.
.

\begin{table}[H]
	\centering
	
	{\centering \par}
	\label{1}
	\begin{tabular}{|c | c| c|}
		\hline
		\textcolor{blue}{Class Number} & \textcolor{blue}{Class Type} & \textcolor{blue}{Number of
			Objects} \\
		\hline
		1 & Mask & 11455 \\
		\hline  
		2 & Improper & 385 \\
		\hline
		3 & No-mask & 8450  \\
		\hline
		4 & Glove & 3175\\
		\hline
		5 & No-glove & 11477\\
		\hline
		
	\end{tabular}
	\caption{Our dataset}
	\label{tab:result2}
\end{table}

\begin{figure}[H] 
	\begin{subfigure}[b]{0.5\linewidth}
		\centering
		\includegraphics[width=0.75\linewidth]{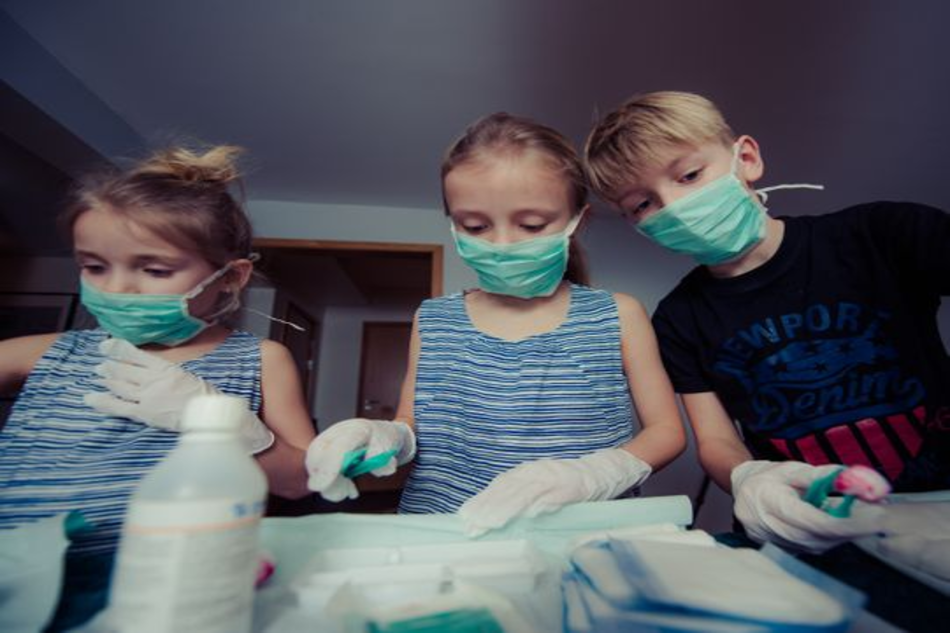} 
		\label{fig7:a} 
		\vspace{4ex}
	\end{subfigure}
	\begin{subfigure}[b]{0.5\linewidth}
		\centering
		\includegraphics[width=0.75\linewidth]{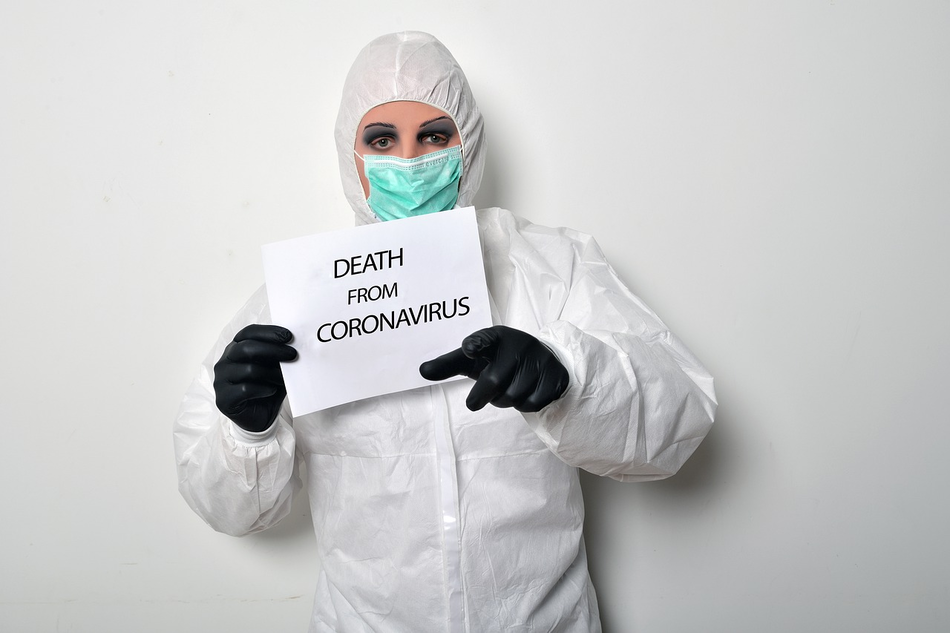} 
	
		\label{fig7:b} 
		\vspace{4ex}
	\end{subfigure} 
	\caption{The ideal group: wearing both masks and gloves. }
	\label{fig7:ideal} 
\end{figure}

Figure \ref{fig7:ideal} shows those who wear protective equipment completely.

\begin{figure}[H] 
	\begin{subfigure}[b]{0.5\linewidth}
		\centering
		\includegraphics[width=0.75\linewidth]{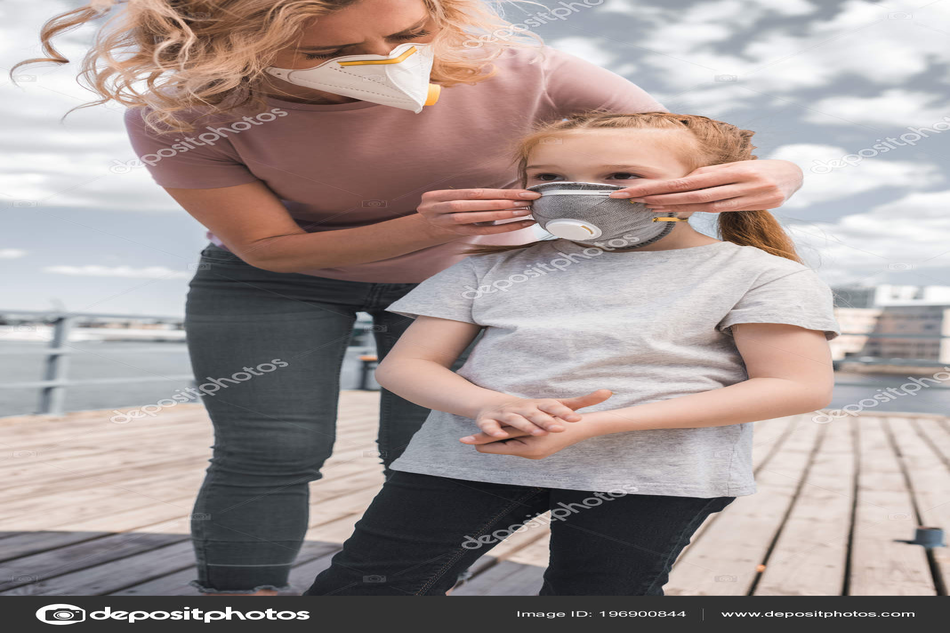} 
		\label{fig7:a} 
		\vspace{4ex}
	\end{subfigure}
	\begin{subfigure}[b]{0.5\linewidth}
		\centering
		\includegraphics[width=0.75\linewidth]{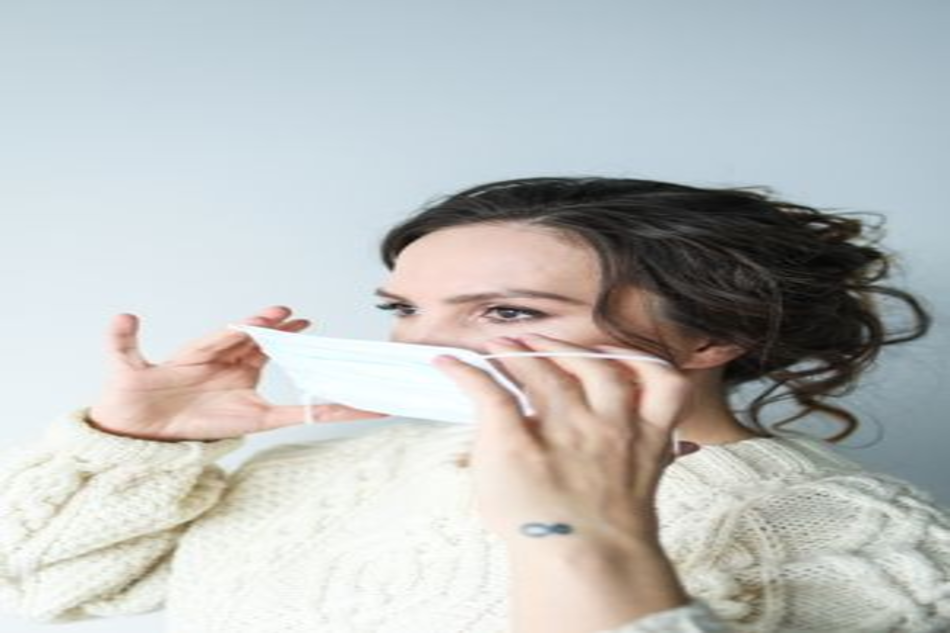} 
	
		\label{fig7:b} 
		\vspace{4ex}
	\end{subfigure}  
	\caption{Although wearing face masks, this group do not wear gloves.}
	\label{fig7:justmask} 
\end{figure}

Figure \ref{fig7:justmask} shows those only wearing masks.

\begin{figure}[H] 
	\begin{subfigure}[b]{0.5\linewidth}
		\centering
		\includegraphics[width=0.75\linewidth]{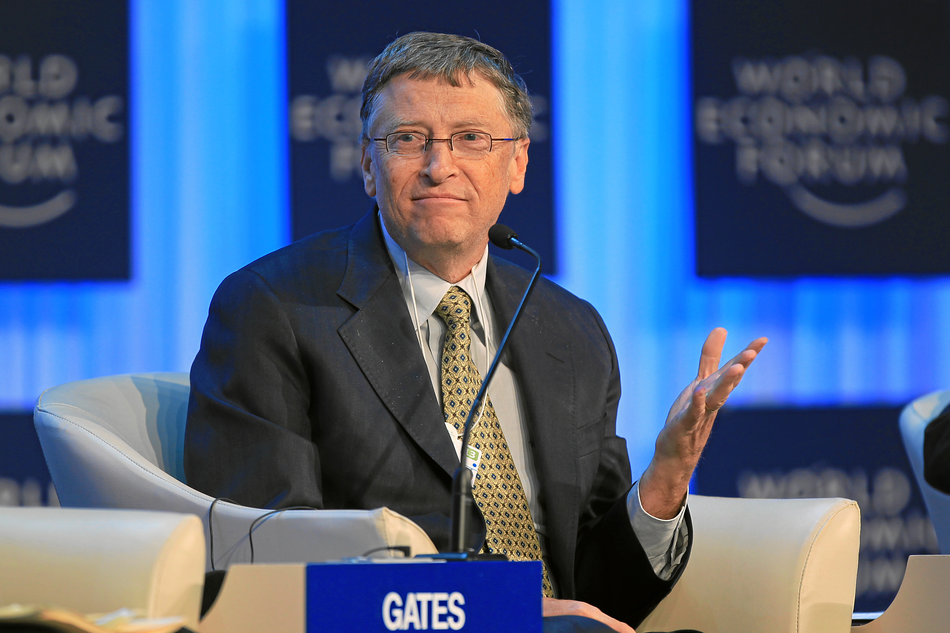} 
	
		\label{fig7:a} 
		\vspace{4ex}
	\end{subfigure}
	\begin{subfigure}[b]{0.5\linewidth}
		\centering
		\includegraphics[width=0.75\linewidth]{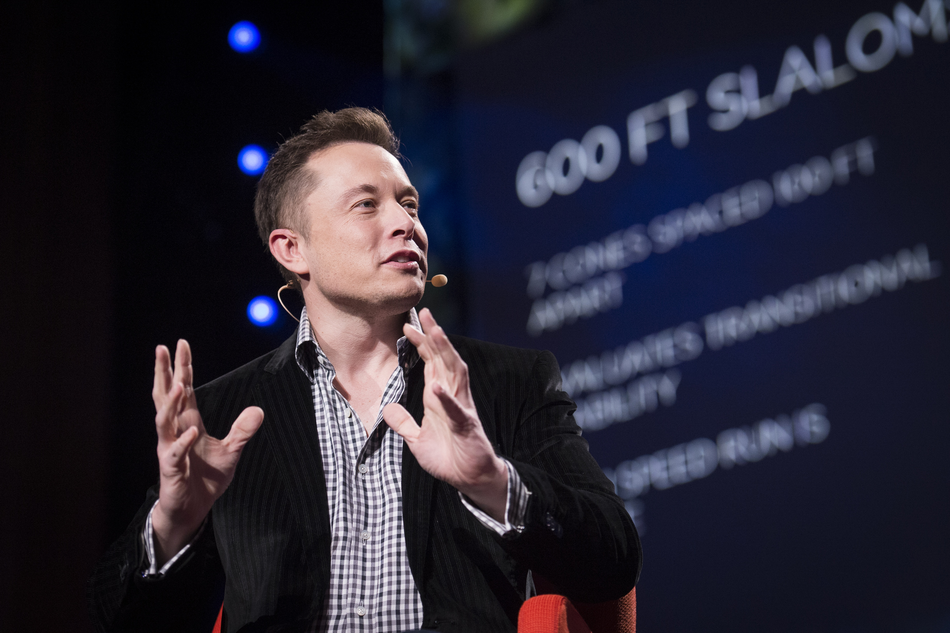} 
		
		\label{fig7:b} 
		\vspace{4ex}
	\end{subfigure} 
	\caption{These images were ideal for making the system more robust so it can detect people that are not serious about the pandemic with high accuracy. }
	\label{fig7:noprotect} 
\end{figure}

\begin{figure}[H] 
	\begin{subfigure}[b]{0.5\linewidth}
		\centering
		\includegraphics[width=0.75\linewidth]{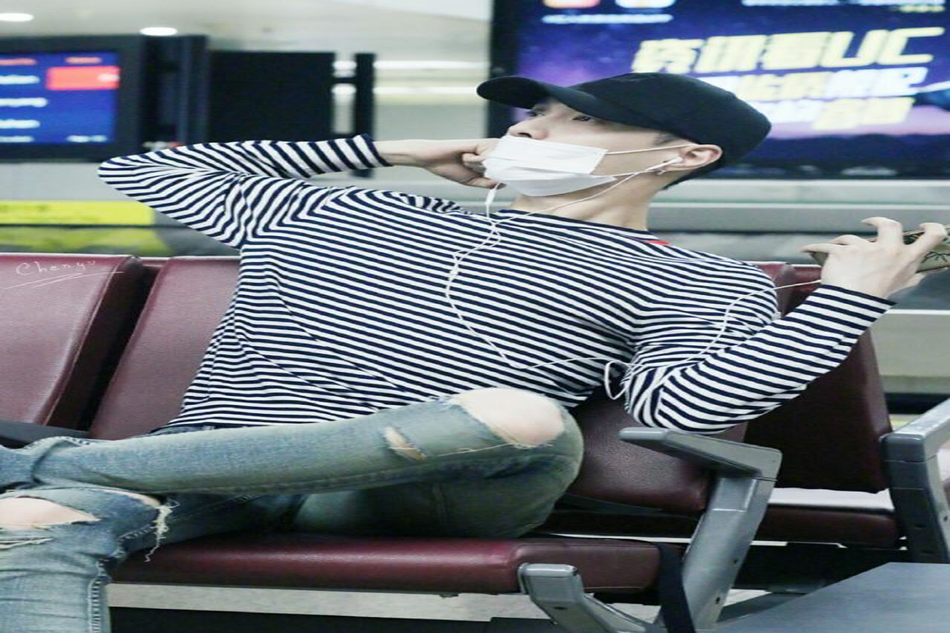} 
		\label{fig7:a} 
		\vspace{4ex}
	\end{subfigure}
		\centering
		\begin{subfigure}[b]{0.5\linewidth}
		\includegraphics[width=0.75\linewidth]{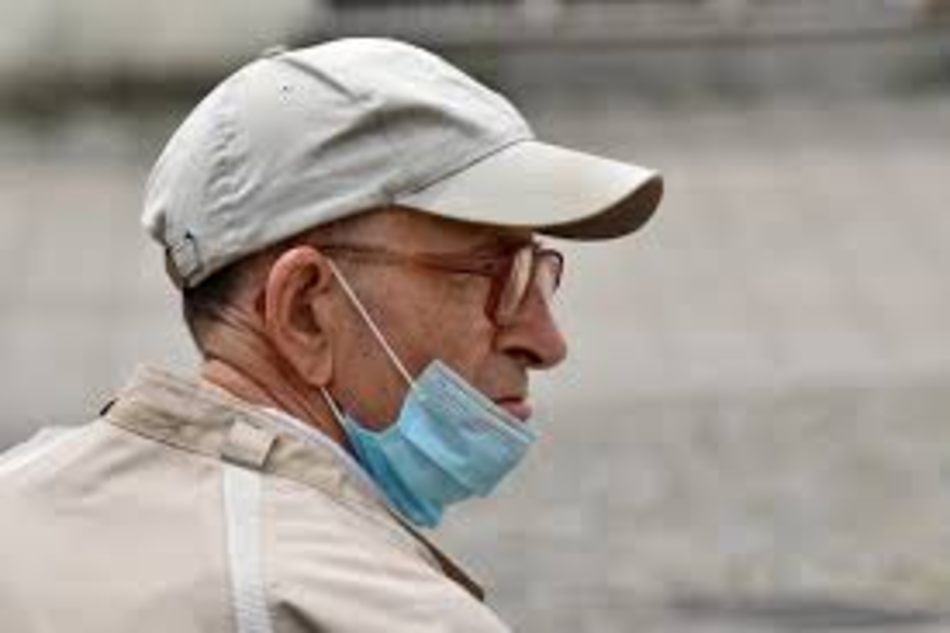} 
		\label{fig7:b} 
		\vspace{4ex}
	\end{subfigure} 
	\caption{Not wearing the mask properly.}
	\label{fig7:improperlywear} 
\end{figure}

Finally, figures \ref{fig7:noprotect} and \ref{fig7:improperlywear} indicate people who do not obey the public rules and those who wear the protections improperly. It is also worth mentioning that some selected images include different classes in the collected dataset.

\subsection{Evaluation measures}
\begin{figure}[H]
	\includegraphics[width=.24\textwidth]{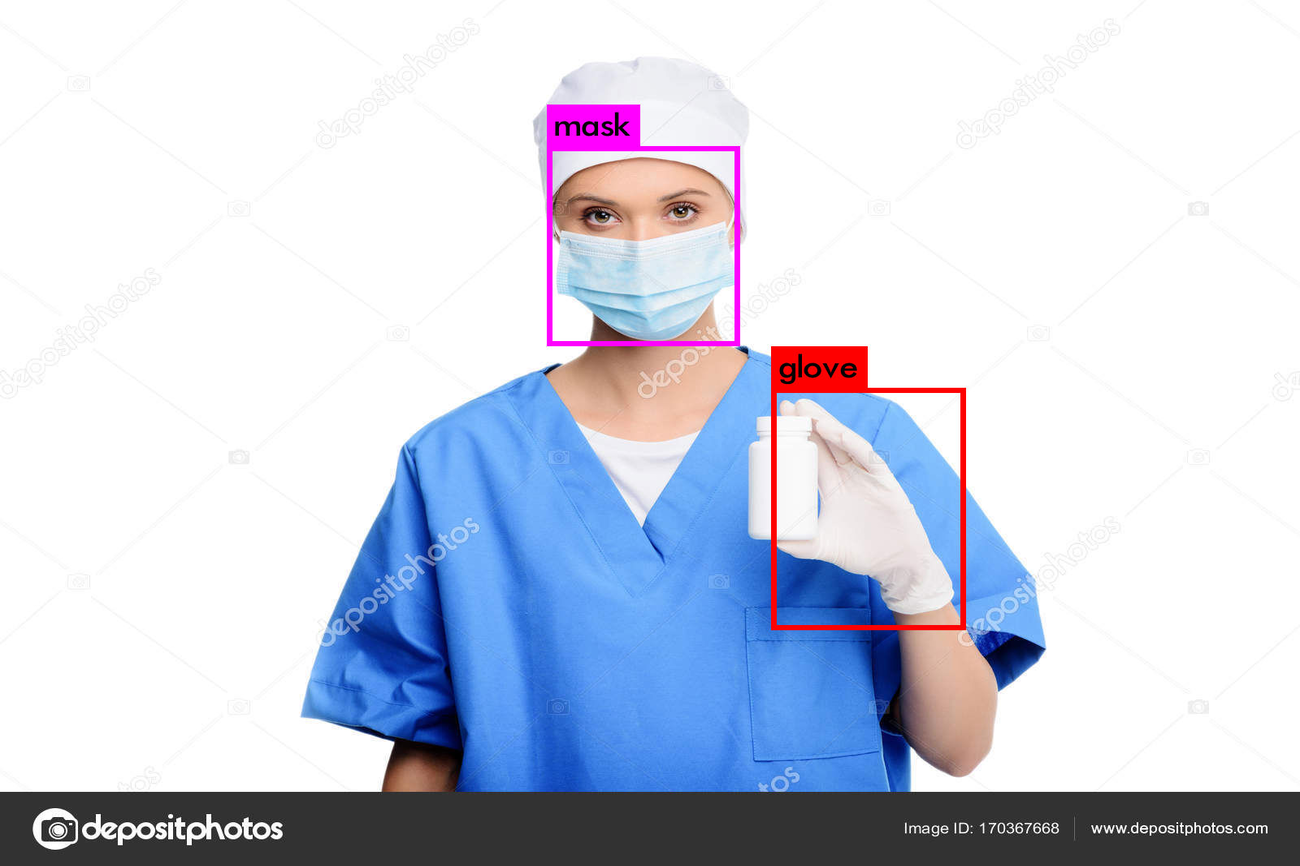}\hfill
	\includegraphics[width=.24\textwidth]{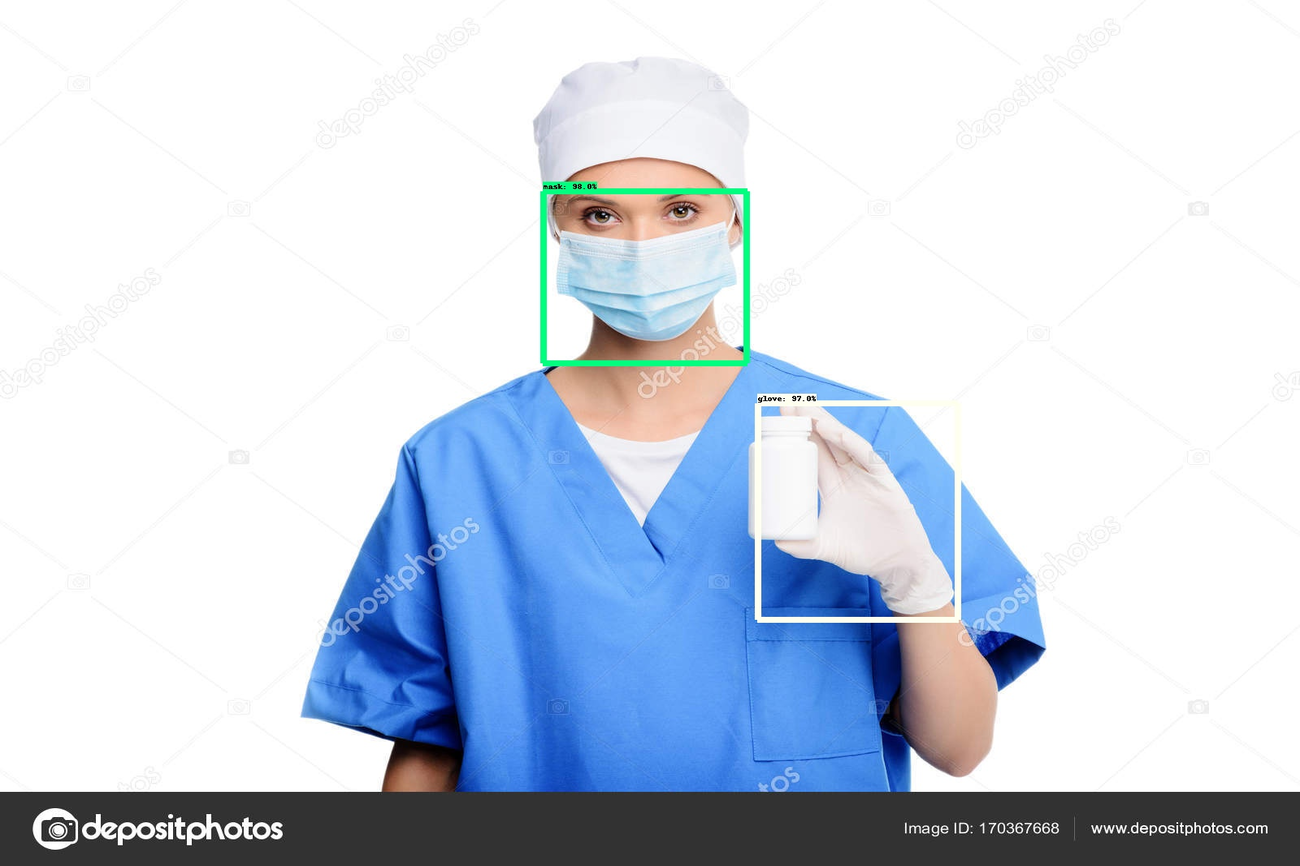}\hfill
	\includegraphics[width=.24\textwidth]{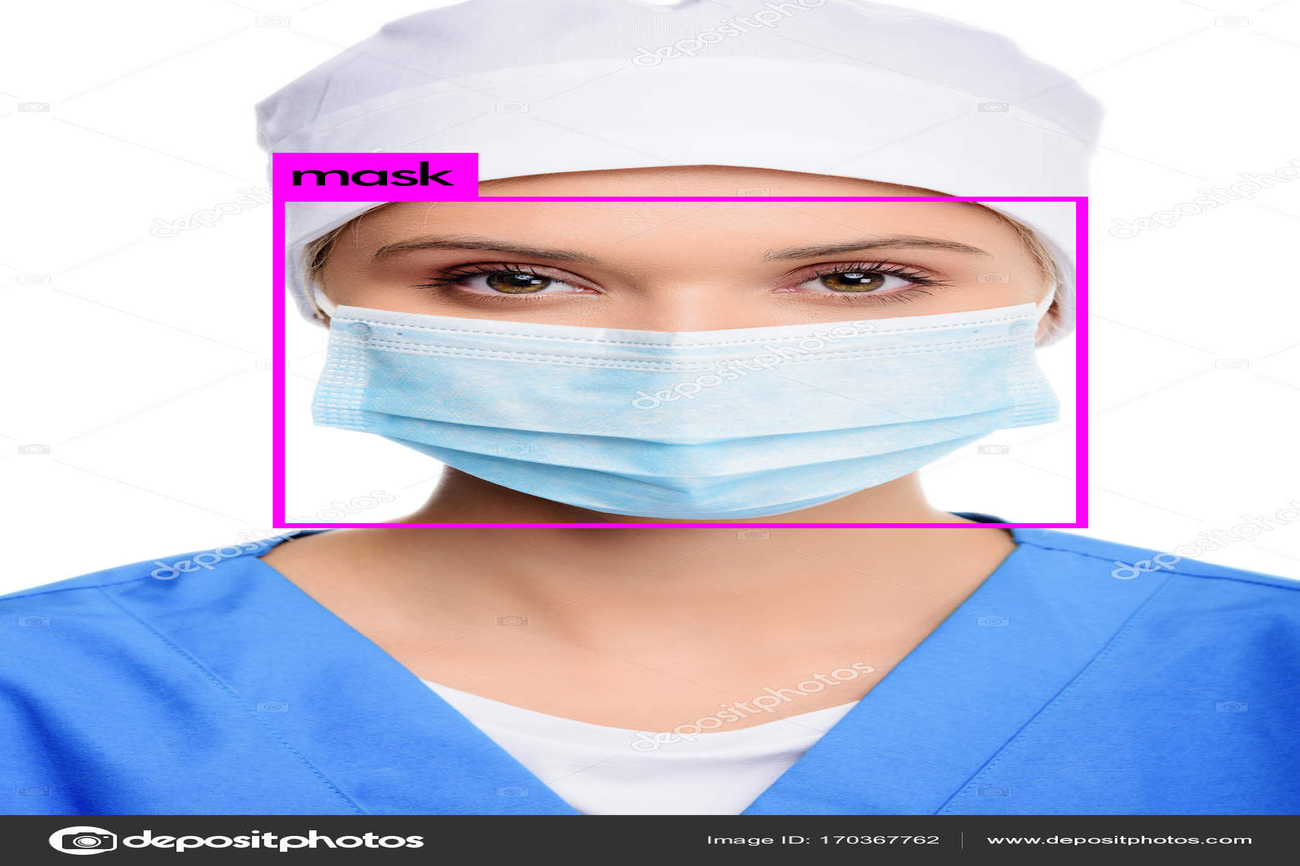}\hfill
	\includegraphics[width=.24\textwidth]{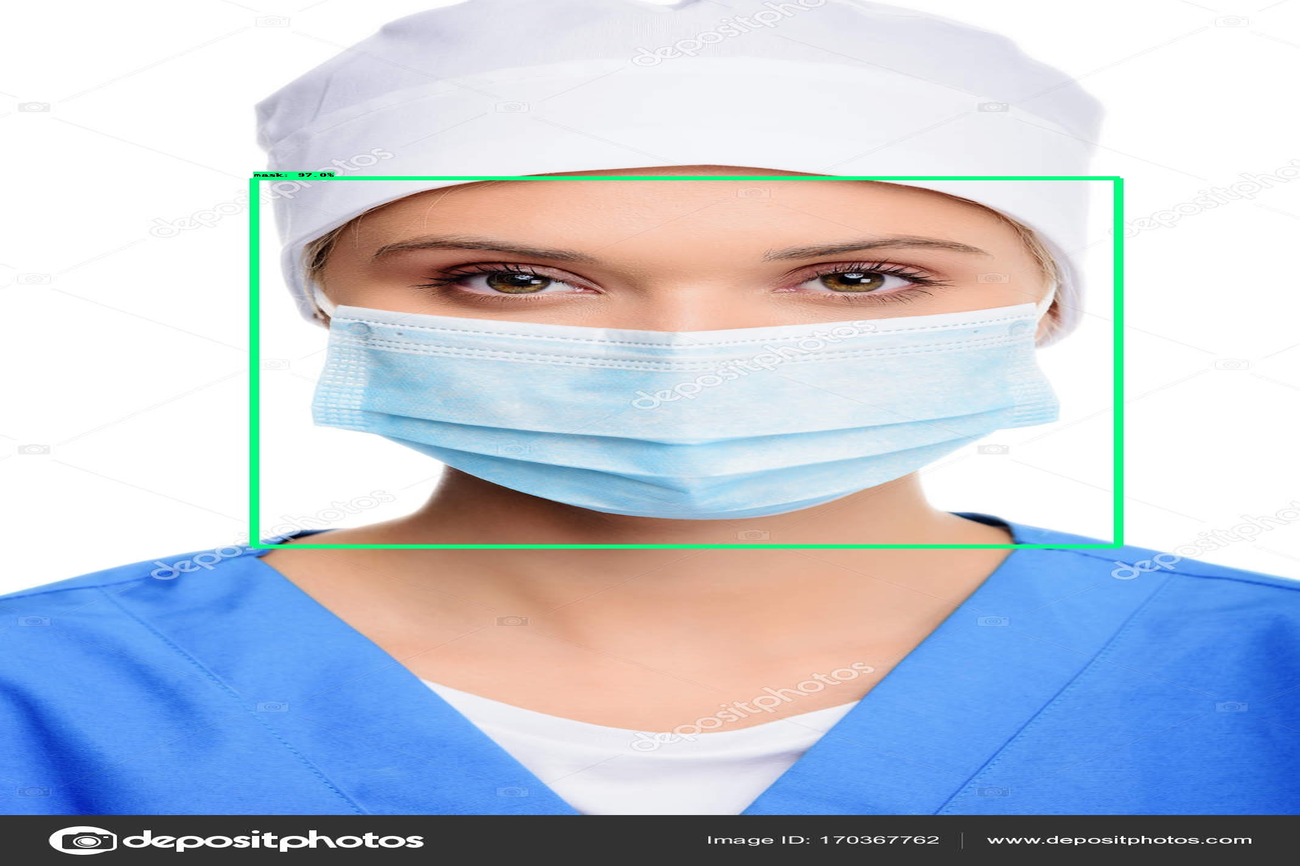}
	\\[\smallskipamount]
	\includegraphics[width=.24\textwidth]{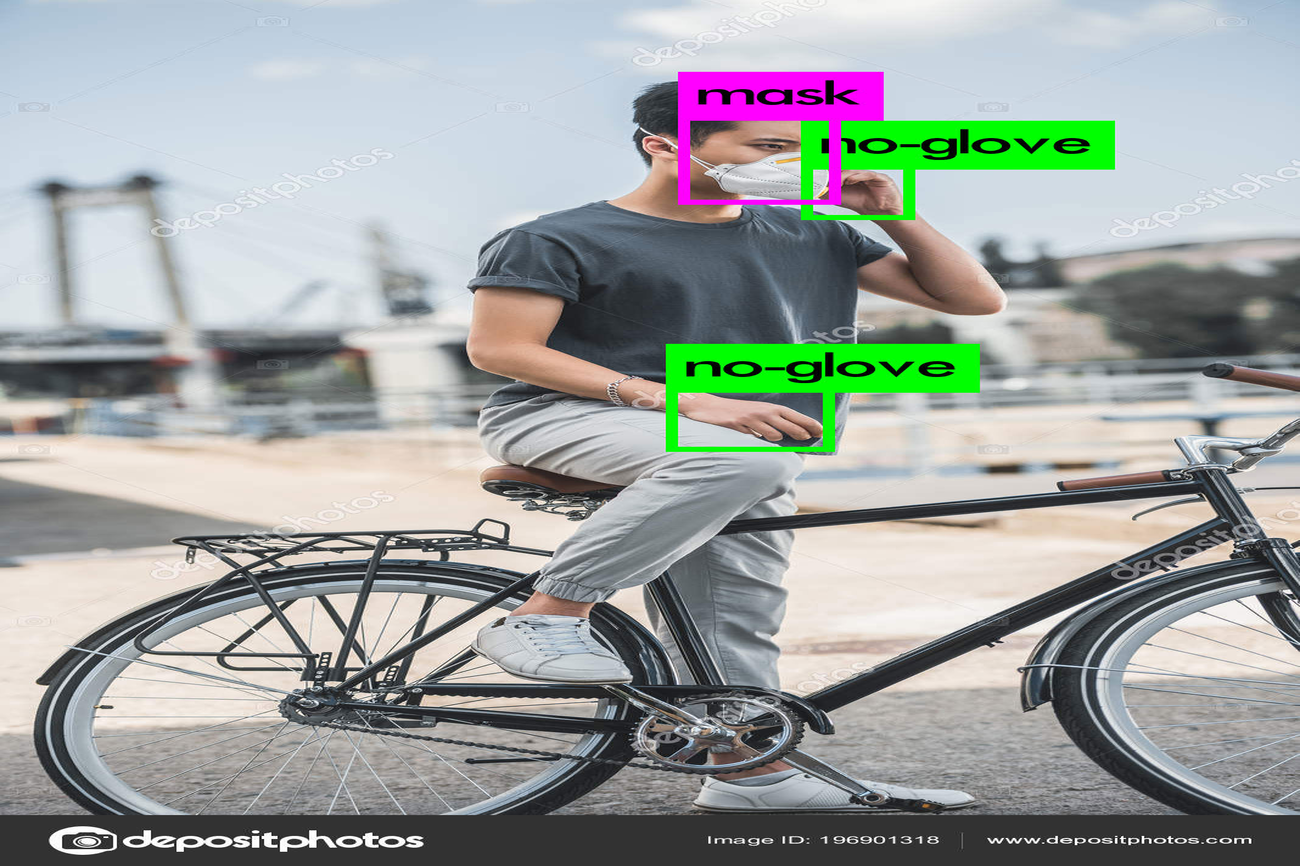}\hfill
	\includegraphics[width=.24\textwidth]{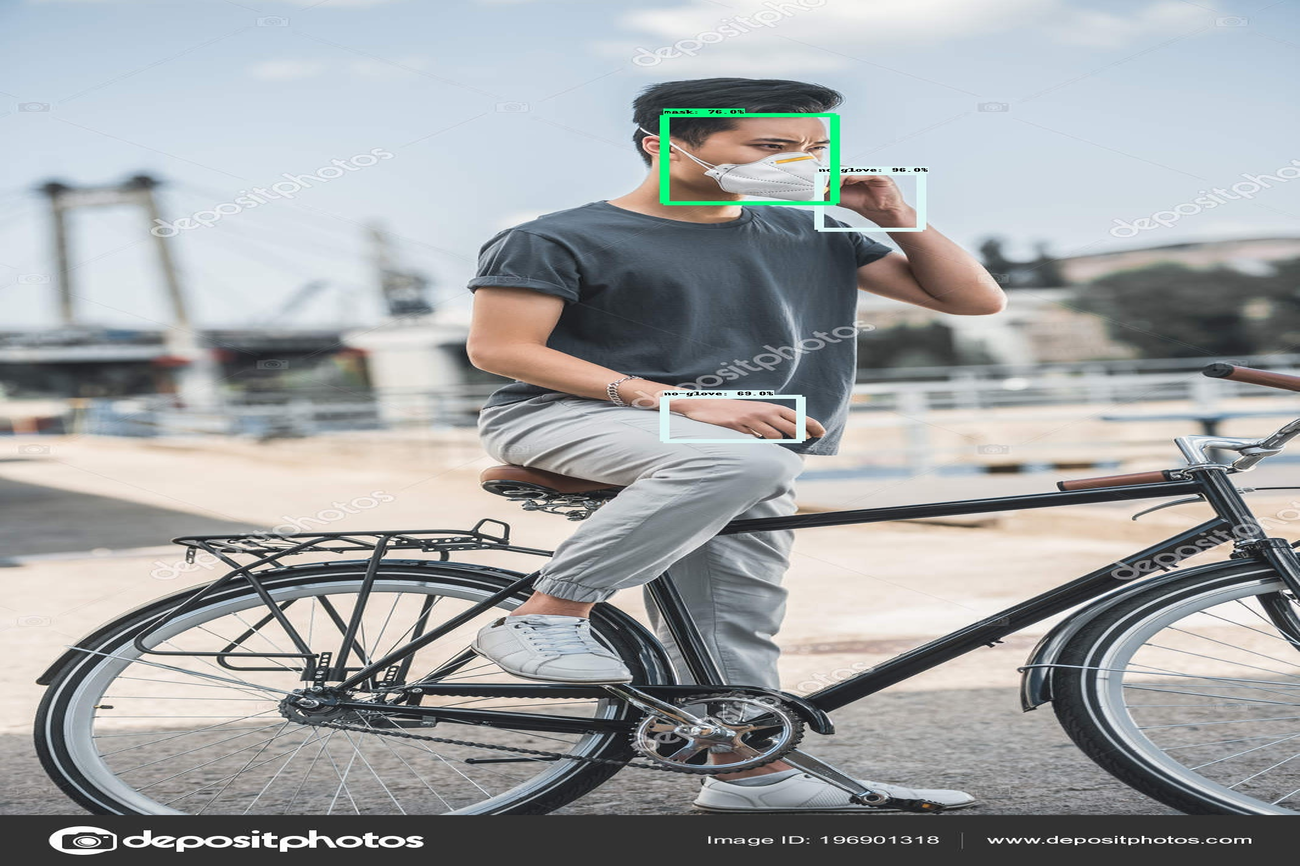}\hfill
	\includegraphics[width=.24\textwidth]{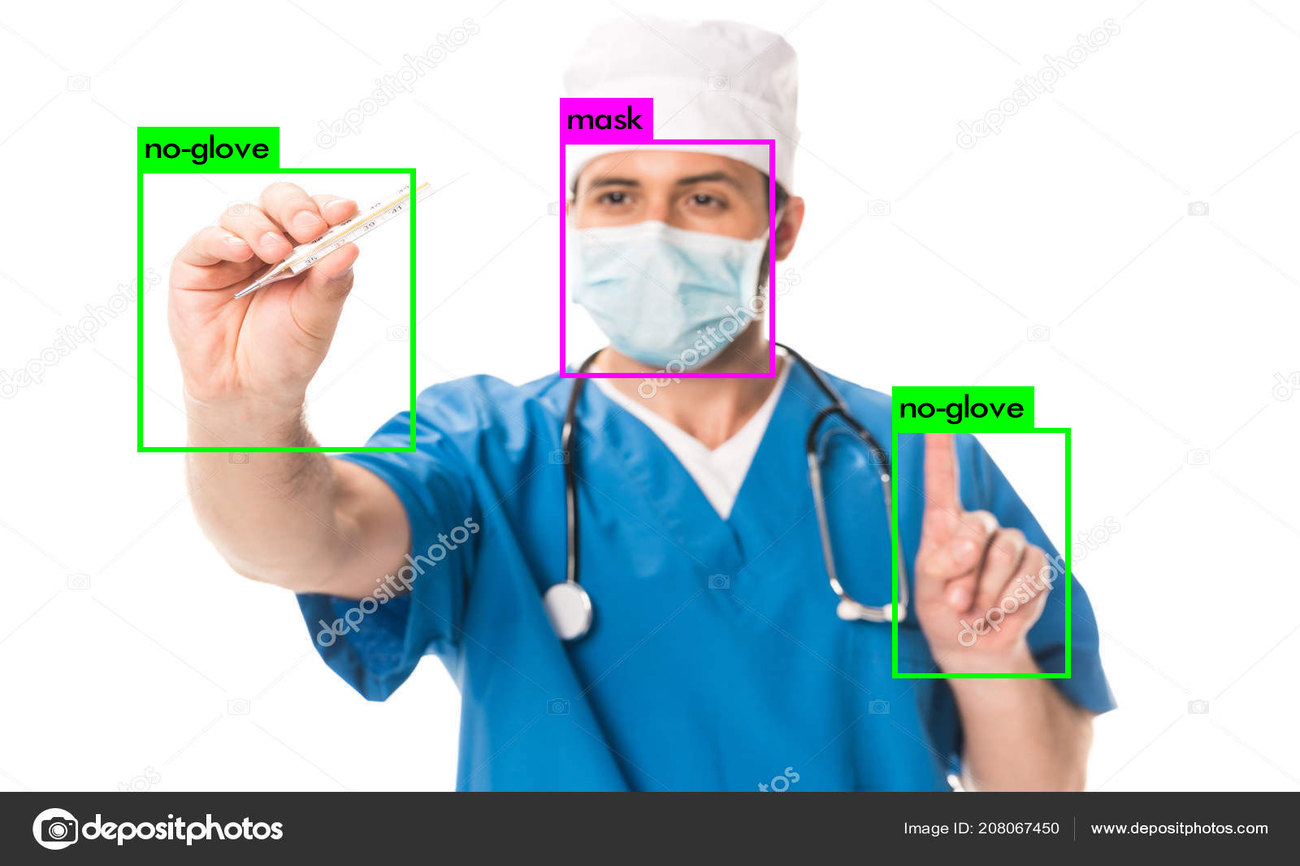}\hfill
	\includegraphics[width=.24\textwidth]{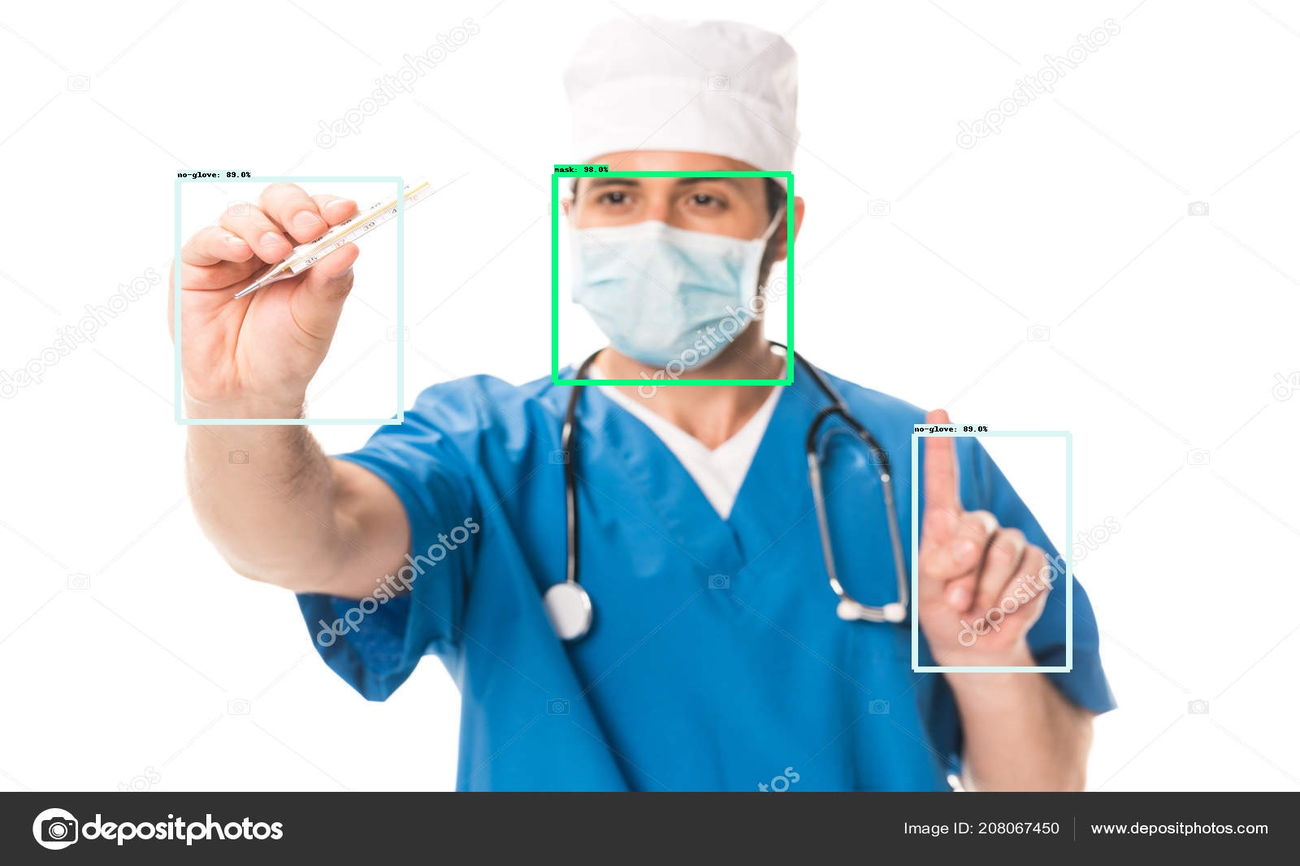}
	\\[\smallskipamount]
	\includegraphics[width=.24\textwidth]{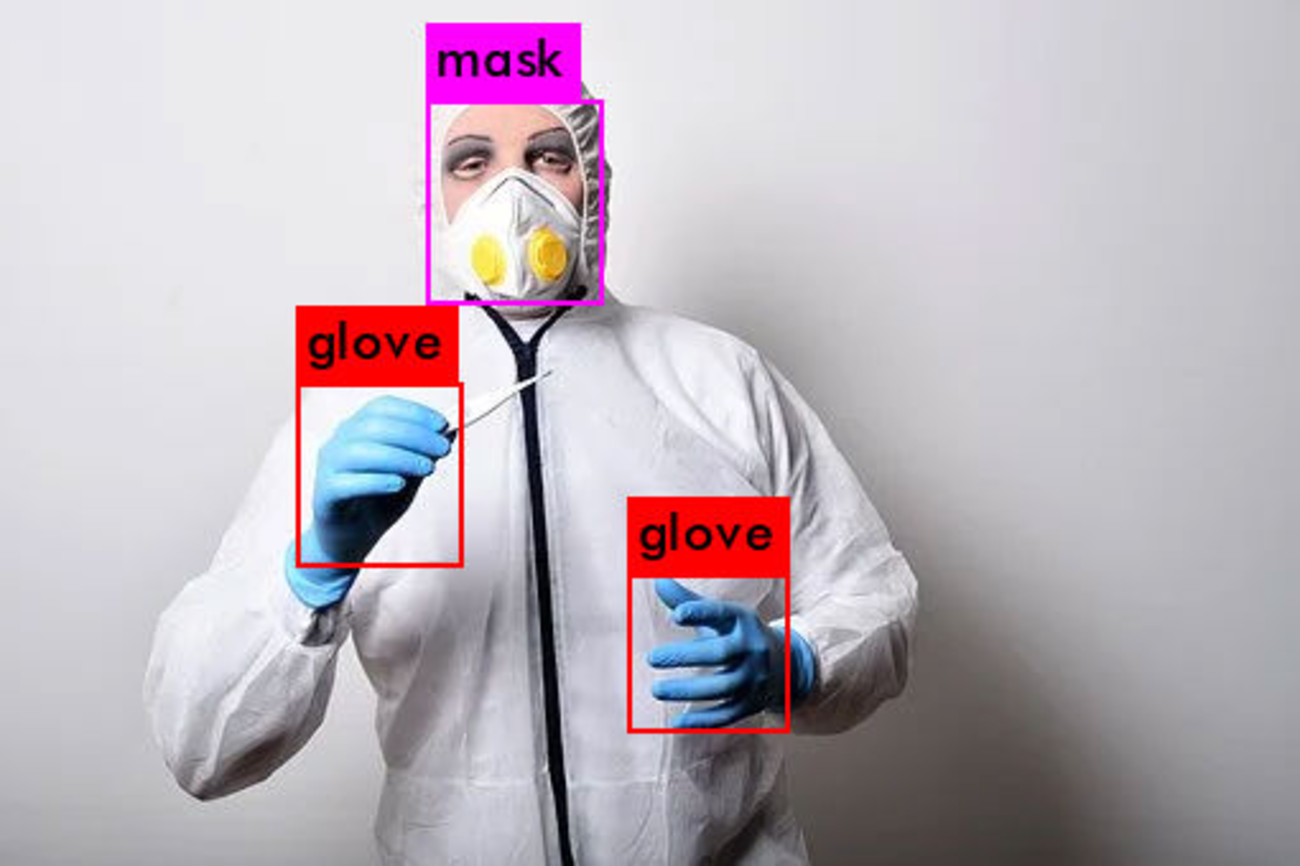}\hfill
	\includegraphics[width=.24\textwidth]{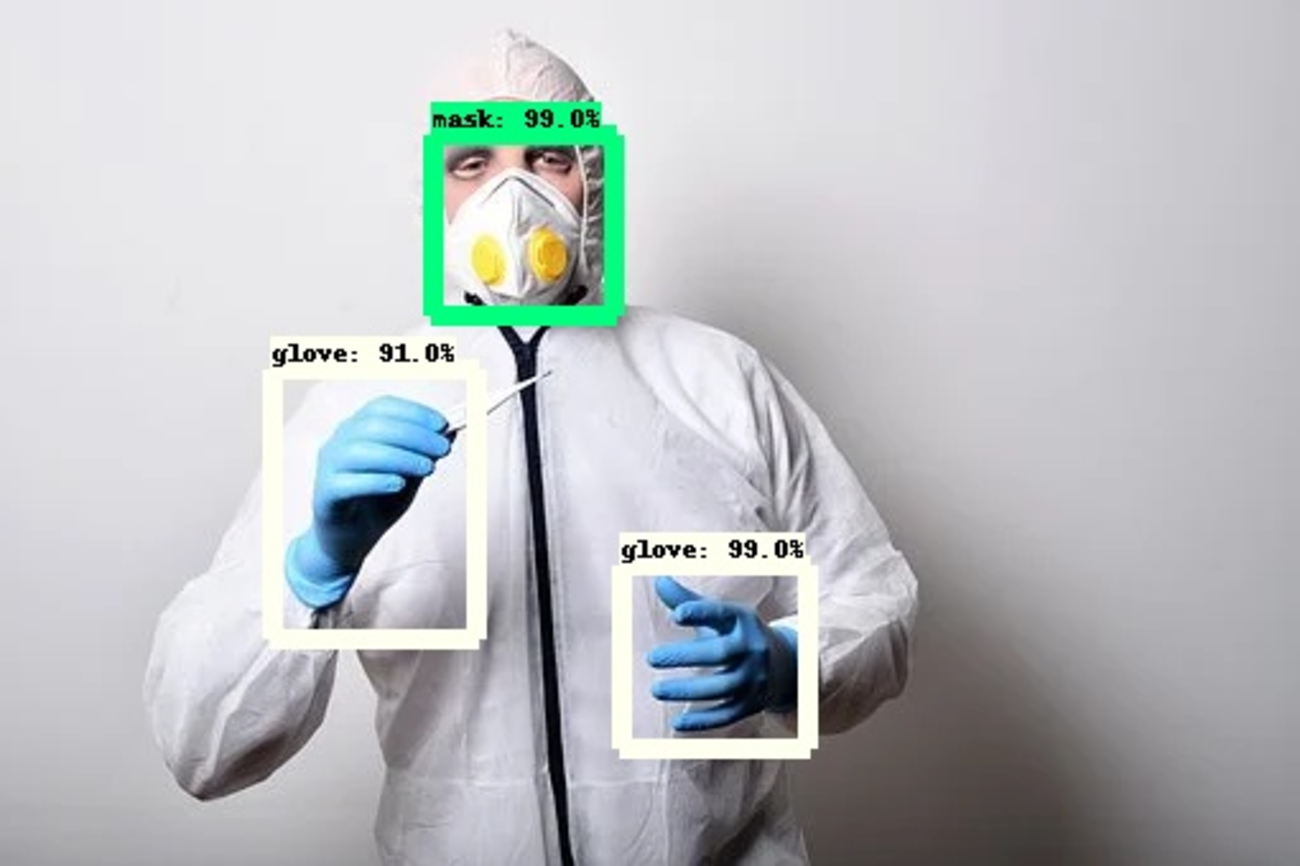}\hfill
	\includegraphics[width=.24\textwidth]{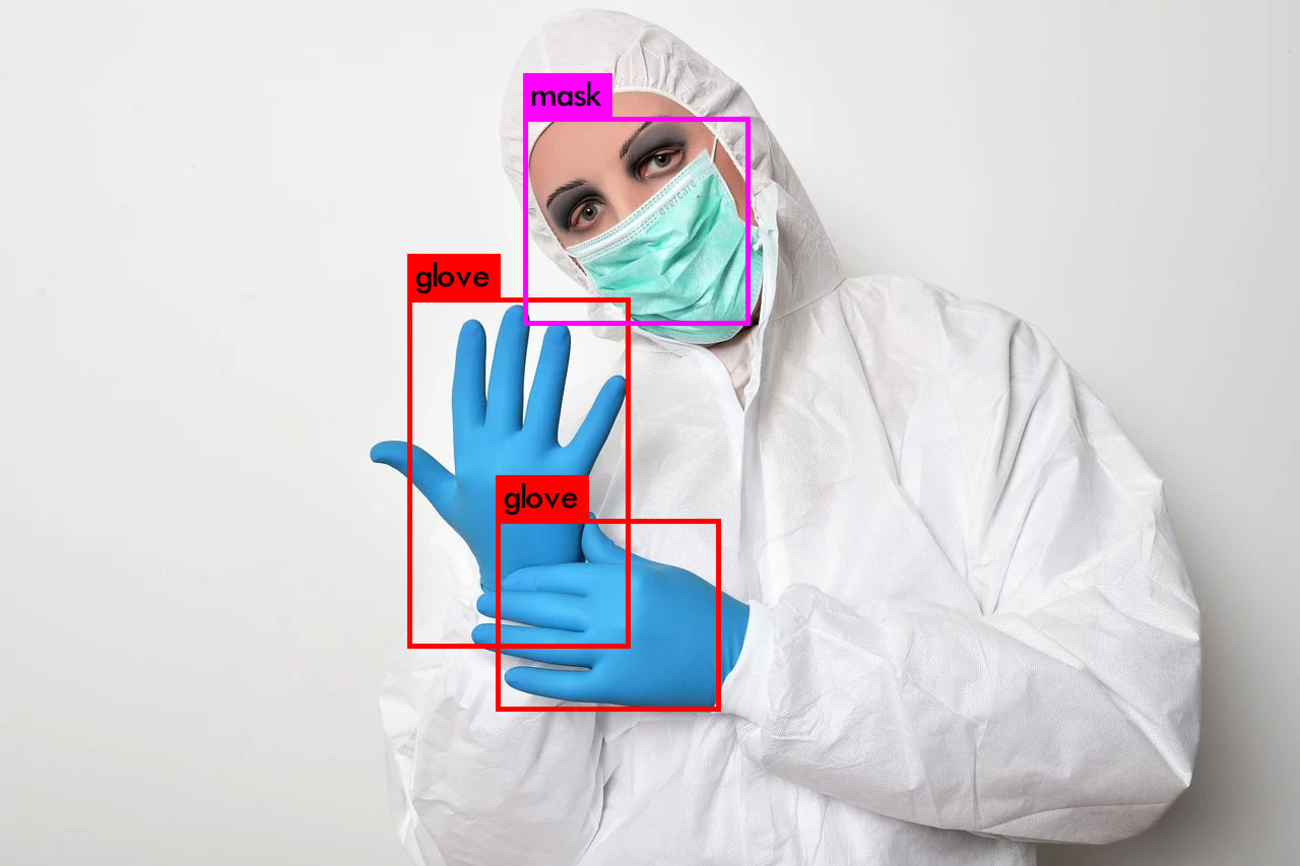}\hfill
	\includegraphics[width=.24\textwidth]{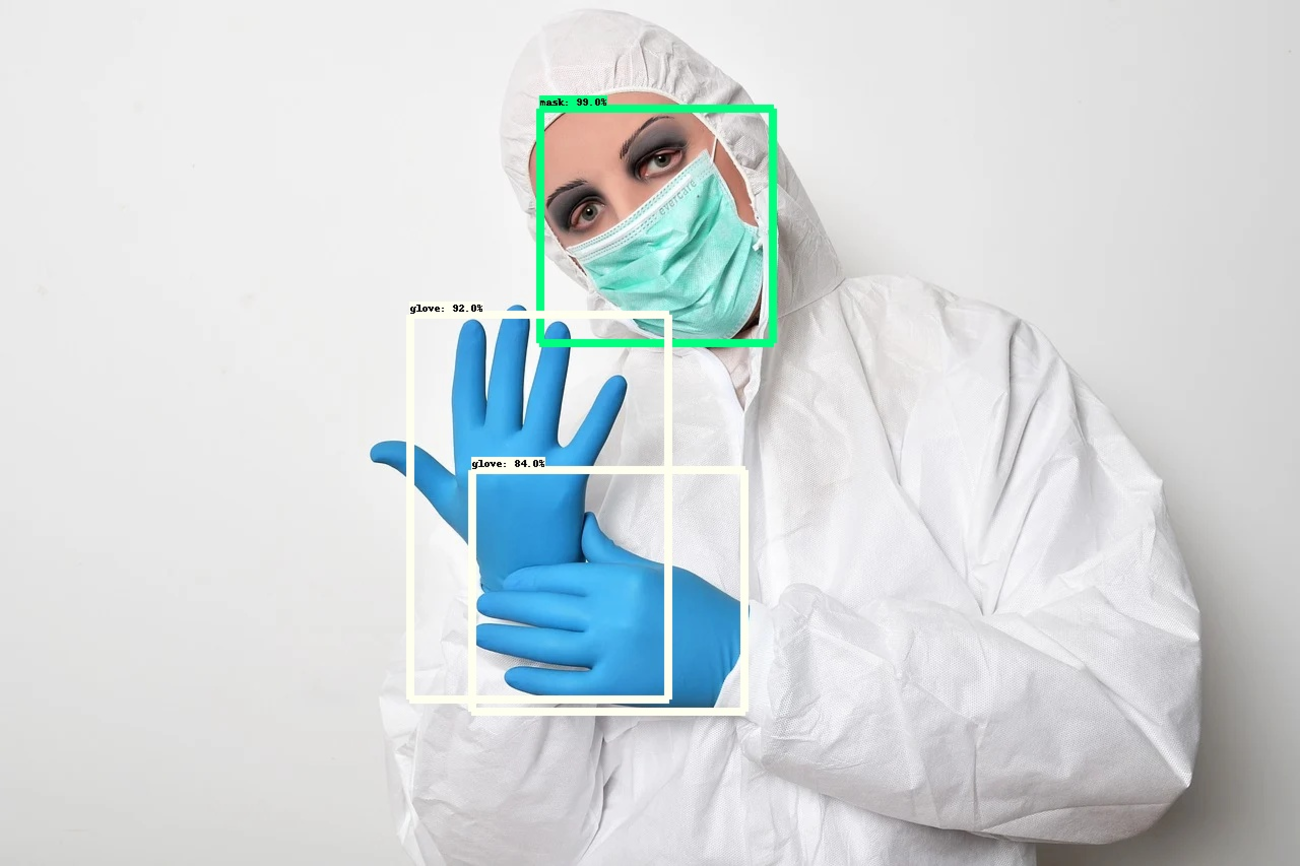}
	\\[\smallskipamount]
	\includegraphics[width=.24\textwidth]{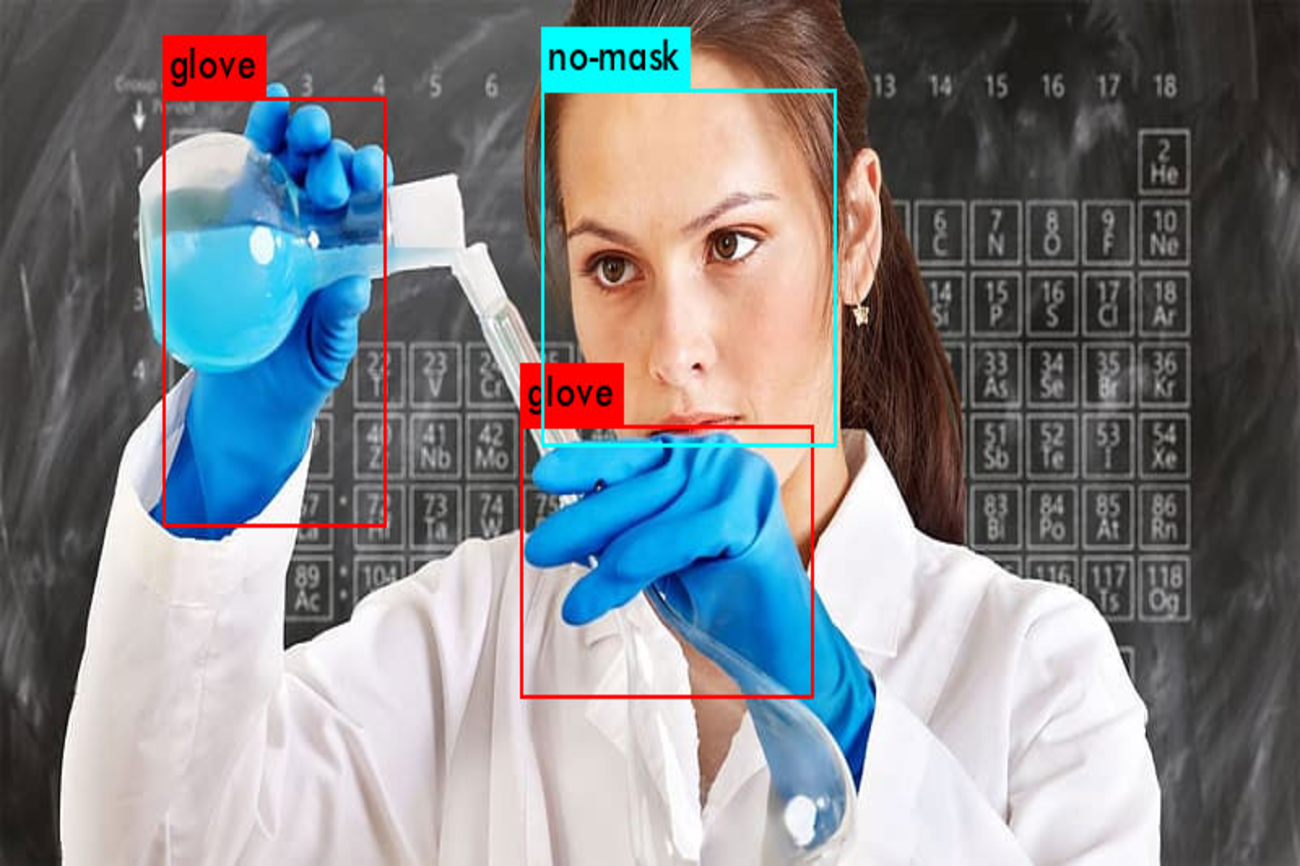}\hfill
	\includegraphics[width=.24\textwidth]{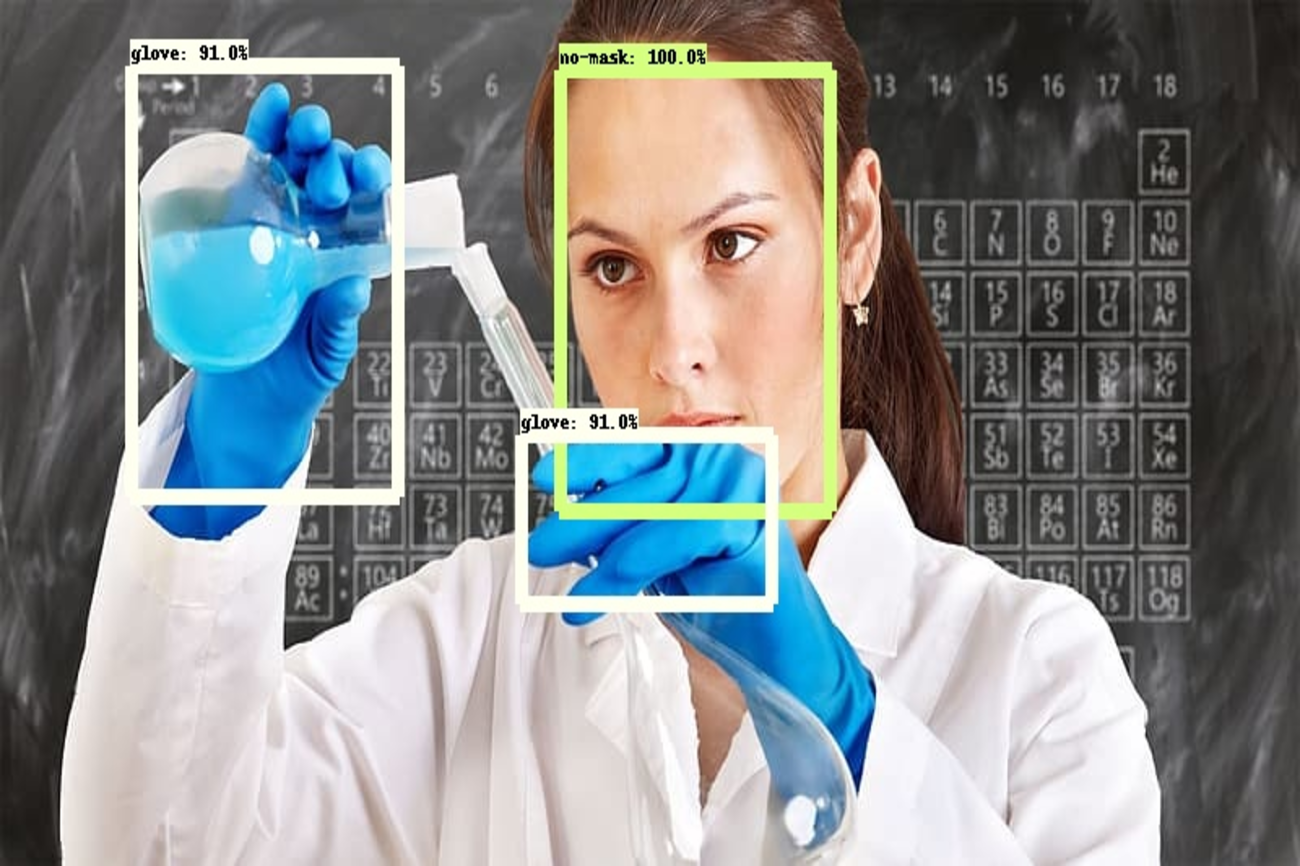}\hfill
	\includegraphics[width=.24\textwidth]{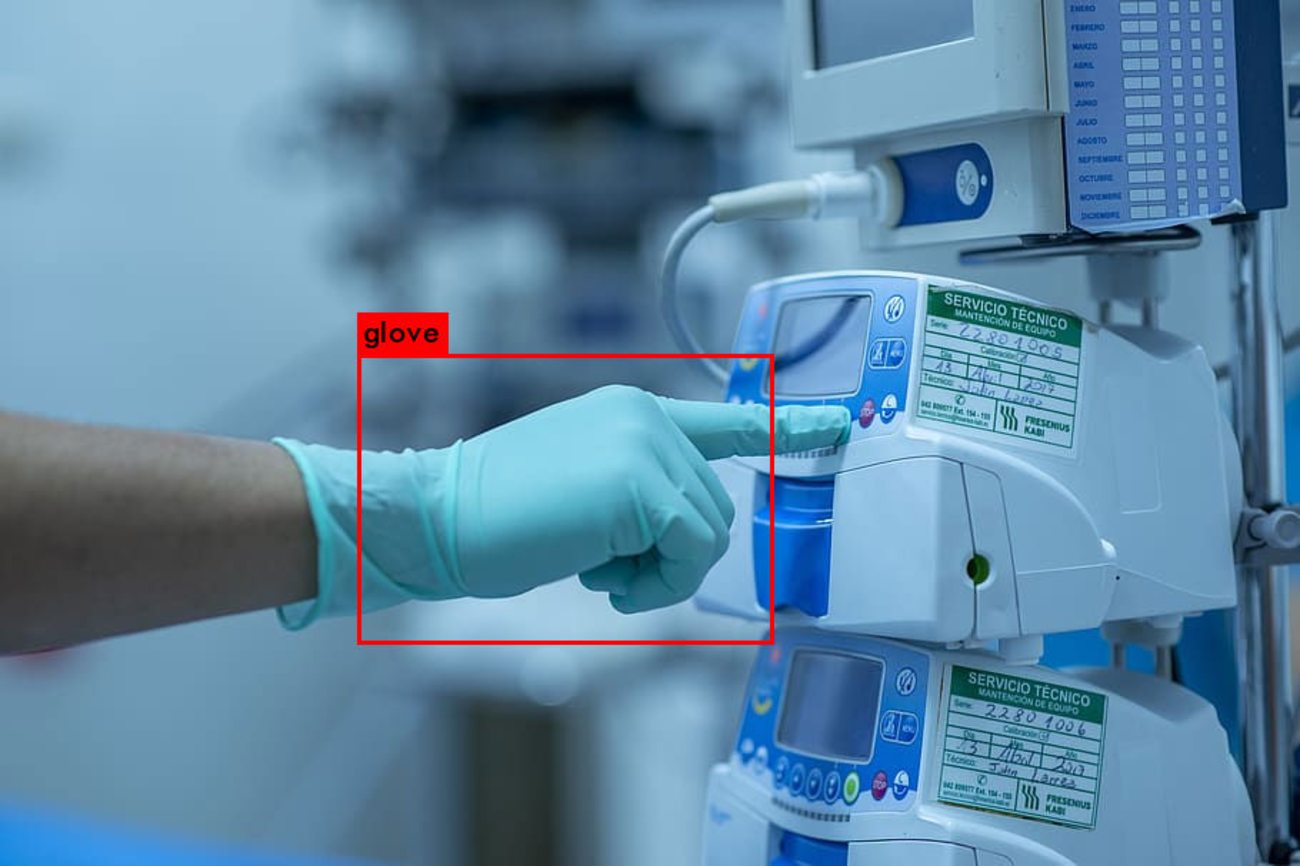}\hfill
	\includegraphics[width=.24\textwidth]{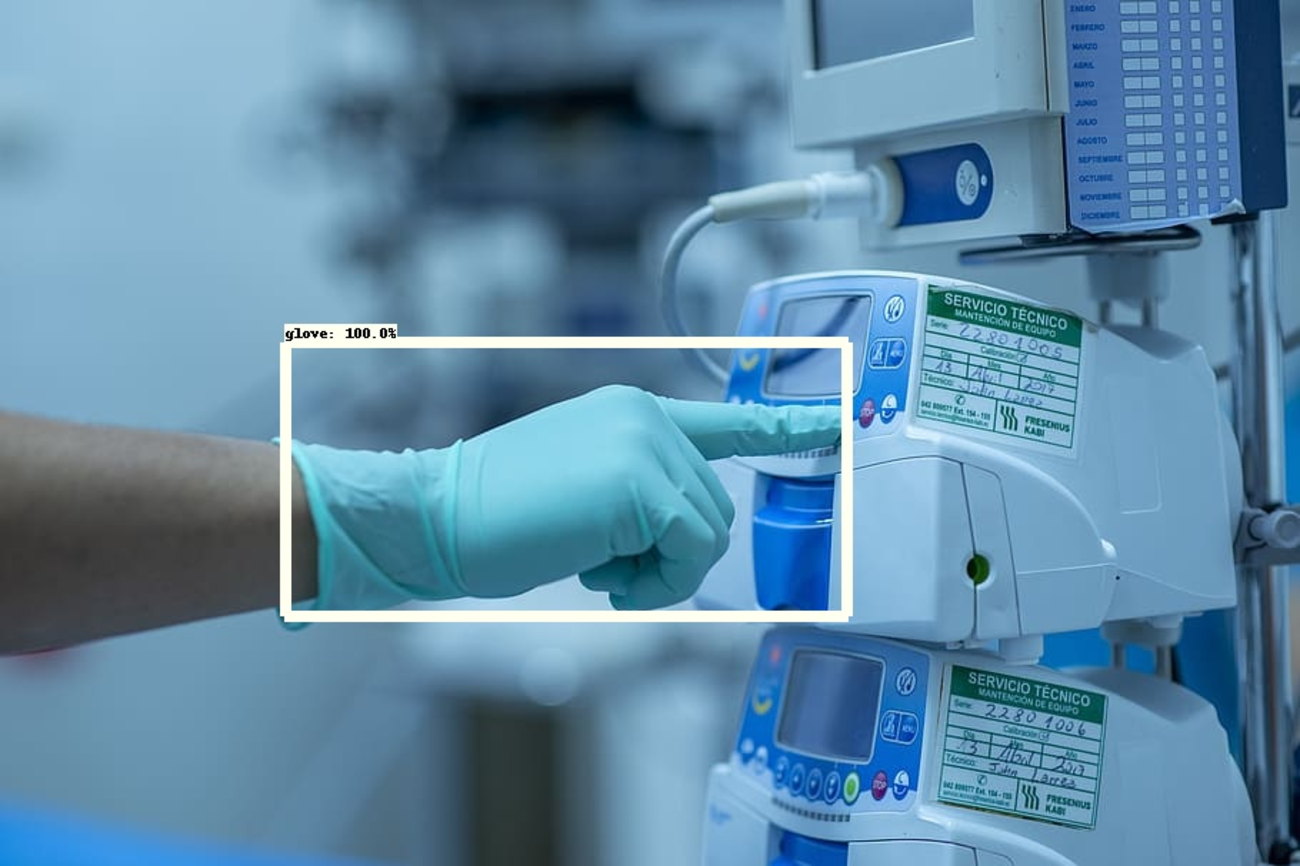}
	\\[\smallskipamount]
	\caption{Image pairs are shown in which the left image is for YOLO, while the
		right image is for SSD. As it can be seen SSD has some weakness in detecting small objects (class: Glove) compared to YOLO.\\ }\label{fig:foobar}

Precision of YOLO and SSD is compared based on object size in table \ref{tab:result}. As it can be seen SSD performs better in detecting bigger objects.
	
\begin{table}[H]
	\centering
	\begin{tabular}{l l l l}
		\hline
		\textbf{method} & \textbf{small} & \textbf{medium}  & \textbf{large}\\
		\hline
		YOLO & \checkmark & \checkmark &   \\
		SSD  &  &  & \checkmark \\
		\hline
	\end{tabular}
	\caption{qualitative comparison based on object size}
	\label{tab:result}
\end{table}
\end{figure}

\begin{figure*}[ht]
	\centering
	\begin{subfigure}[b]{0.475\textwidth}
		\centering
		\includegraphics[width=\textwidth]{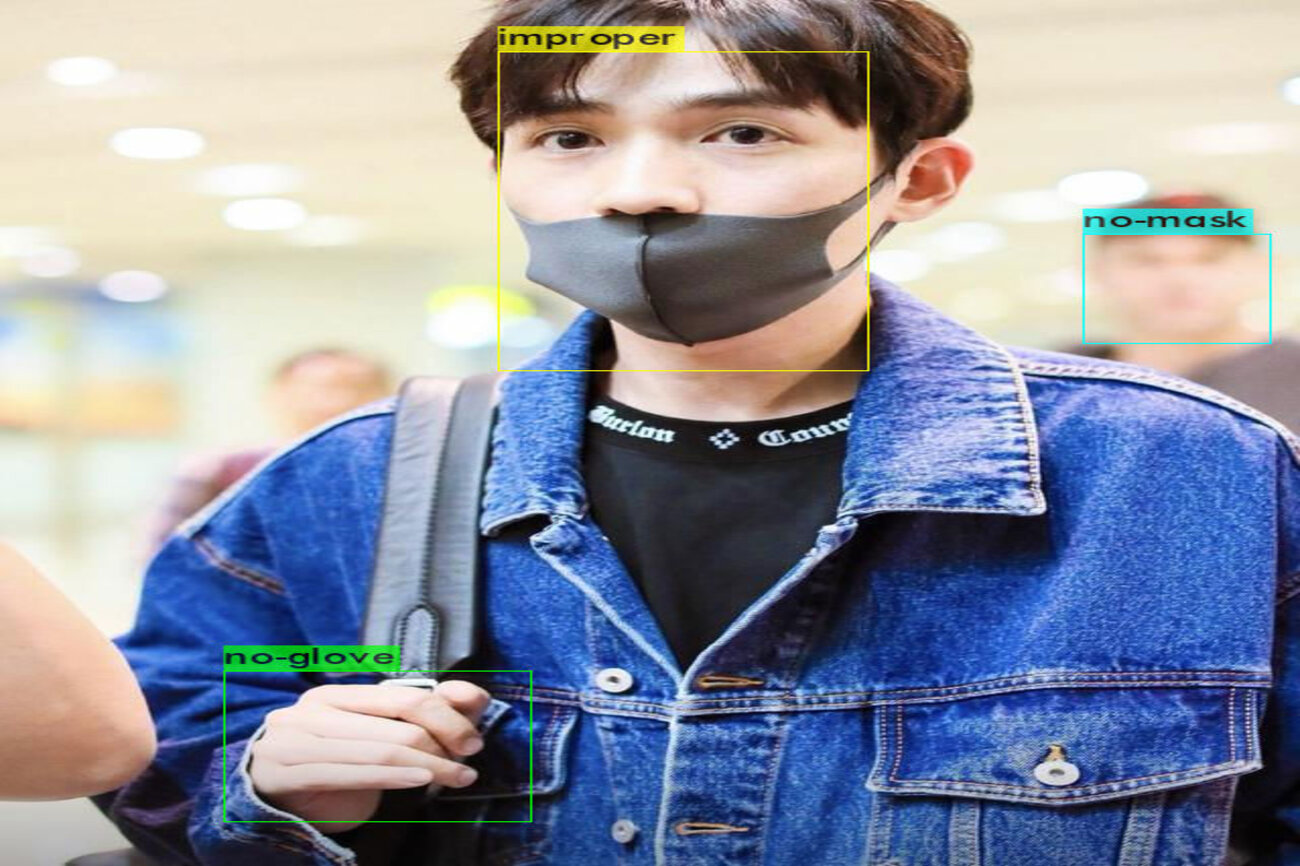}
		\caption[]%
		{{\small Correctly classified by YOLO}}    
		\label{fig:shapebased}
	\end{subfigure}
	\hfill
	\begin{subfigure}[b]{0.475\textwidth}  
		\centering 
		\includegraphics[width=\textwidth]{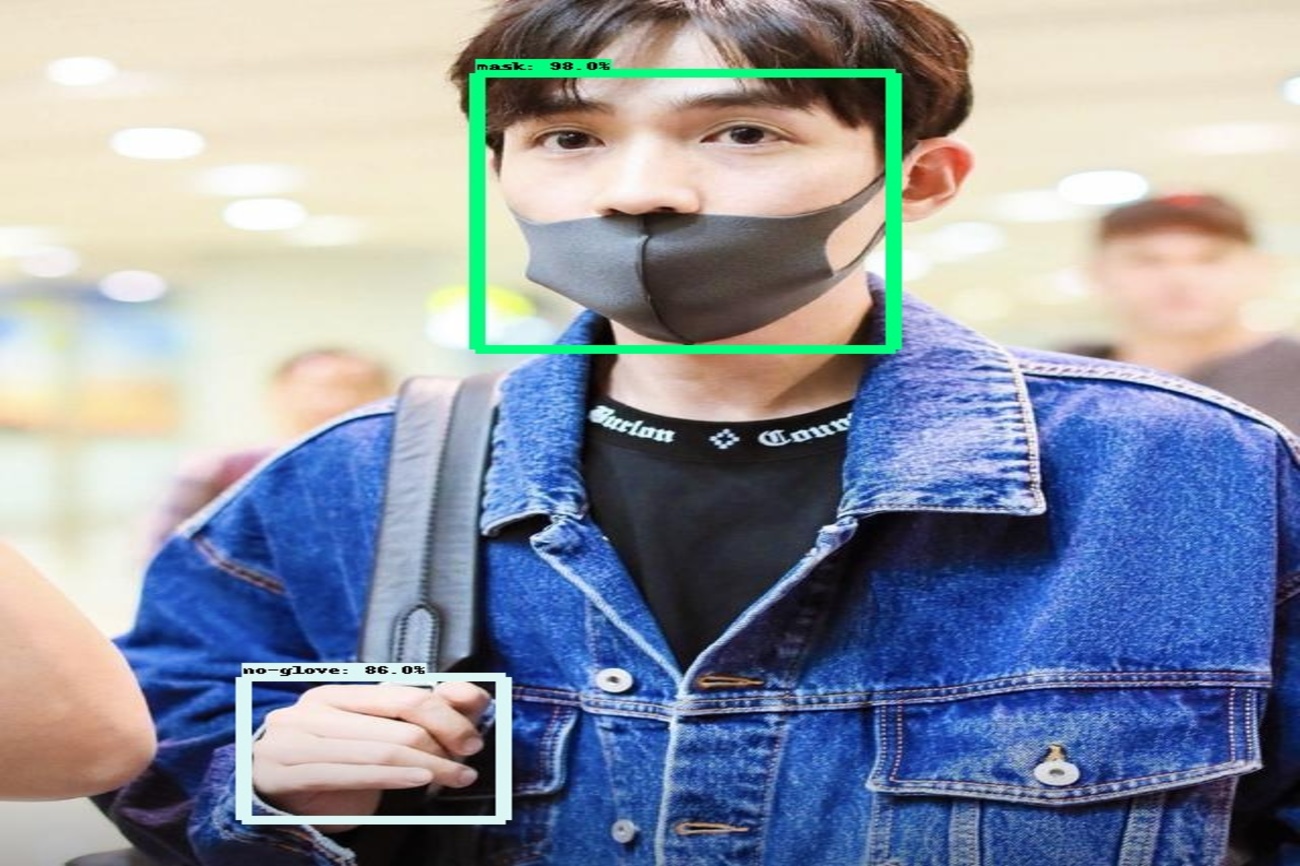}
		\caption[]%
		{{\small Misclassified by SSD}}    
		\label{fig:recover}
	\end{subfigure}
	\caption[  ]
	{\small b is one of misclassified examples in our dataset which come from the high similarity in No-mask and Improper classes.} 
	\label{fig:mis}
\end{figure*}

mAP is a metric used for evaluating object detectors. It is the average of the AP, first precision and recall are defined for a single data for better understanding.\\
PRECISION = $\dfrac{TP}{TP+FP}$ $\qquad$
RECALL = $\dfrac{TP}{TP+FN}$
\begin{itemize}
	\item TP = A correct detection and IOU is greater than or equal to the
	threshold. 
	\item FP = A wrong detection or IOU
	is less than the threshold.
	\item FN = no detection for ground truth.
\end{itemize}
Precision and recall are always between 0 and 1, where AP is defined as the area under the precision-recall curve, which is a plot of precision as a function of recall.$$AP = \int_{0}^{1} precision(r) dr$$
We compute the AP for each class and average them at different IOU values. However, mAp@(0.5), which is the average of AP over  IOU=0.5, means that 50\% or greater overlap of the detected box with its ground truth would be considered true positive is the main metric for results. The results of object detection using YOLO and SSD methods are depicted in table \ref{tab:yoloresult} and \ref{tab:ssdresult} where Area determines the size of objects and mAp@[.5:.95] means the average mAp over different IOU thresholds, from 0.5 to 0.95 by step of 0.05 (0.5, 0.55, 0.6, 0.65, 0.7, 0.75, 0.8, 0.85, 0.9, 0.95). AR is the maximum recall given a fixed number of detections per image, which is determined by maxDets. Figure \ref{fig:foobar} shows the results on the data in the collected dataset. Furthermore, Figure \ref{fig:mis} shows some misclassified results. 
\newpage
\subsection{Result tables}
\begin{table}[H]
	\centering
	\begin{tabular}{l l l l}
		\hline
				\textbf{\textcolor{blue}{Mean Average Precision}} & \textbf{\textcolor{blue}{IOU}} & \textbf{\textcolor{blue}{Area}} & \textbf{\textcolor{blue}{maxDets}}\\
		\hline
		0.462 & 0.50:0.95 & all &  100 \\
		\textcolor{red}{0.906} & 0.50 & all &  100 \\
		0.320 & 0.75 & all &  100 \\
		0.593 & 0.50:0.95 & small &  100 \\
		0.385 & 0.50:0.95 & medium &  100 \\
		0.479 & 0.50:0.95 & large &  100 \\
		\hline
			\textbf{\textcolor{blue}{Mean Average Recall}} & \textbf{\textcolor{blue}{IOU}} & \textbf{\textcolor{blue}{Area}} & \textbf{\textcolor{blue}{maxDets}}\\
		\hline
		0.397 & 0.50:0.95 & all &  1 \\
		0.485 & 0.50:0.95 & all &  10 \\
		0.493 & 0.50:0.95 & all &  100 \\
		0.610 & 0.50:0.95 & small &  100 \\
		0.427 & 0.50:0.95 & medium &  100 \\
		0.525 & 0.50:0.95 & large &  100 \\
		\hline
	\end{tabular}
	\caption{YOLO result on our dataset}
	\label{tab:yoloresult}
\end{table}
\begin{table}[H]
	\centering
	\begin{tabular}{l l l l}
		\hline
		\textbf{\textcolor{blue}{Mean Average Precision}} & \textbf{\textcolor{blue}{IOU}} & \textbf{\textcolor{blue}{Area}} & \textbf{\textcolor{blue}{maxDets}}\\
		\hline
		0.521 & 0.50:0.95 & all &  100 \\
		\textcolor{red}{0.855} & 0.50 & all &  100 \\
		0.524 & 0.75 & all &  100 \\
		0.514 & 0.50:0.95 & small &  100 \\
		0.337 & 0.50:0.95 & medium &  100 \\
		0.545 & 0.50:0.95 & large &  100 \\
		\hline
			\textbf{\textcolor{blue}{Mean Average Recall}} & \textbf{\textcolor{blue}{IOU}} & \textbf{\textcolor{blue}{Area}} & \textbf{\textcolor{blue}{maxDets}}\\
		\hline
		0.429 & 0.50:0.95 & all &  1 \\
		0.573 & 0.50:0.95 & all &  10 \\
		0.586 & 0.50:0.95 & all &  100 \\
		0.553 & 0.50:0.95 & small &  100 \\
		0.446 & 0.50:0.95 & medium &  100 \\
		0.594 & 0.50:0.95 & large &  100 \\
		\hline
	\end{tabular}
	\caption{SSD result on our dataset}
	\label{tab:ssdresult}
\end{table}

\newpage
\section{Conclusion}
In this study, two different DNN object detection algorithms have been applied for proper masked face and glove detection. We have compared two popular DDNs: SSD MobilNet and YOLOv3. Both of them were trained via transfer learning. The number of training iteration is selected based on the minimum loss. The result indicates that YOLO outperforms SSD MobileNet in terms of mAp. However, when the average recall is considered, SSD shows a better result. Also, as observed in tables 3 and 4, SSD obtained a better result in IOU=0.75 compared to YOLO, demonstrating better alignment with the target. Considering the situations in public places, better alignment is not a crucial factor in this problem. Whereas, the ability of a method to detect objects with acceptable accuracy is more critical even with less alignment. So, YOLOv3 may be a more useful detection method. However, SSD might perform better in low computational power systems since of a lighter architecture than YOLO. This work has the potential to operate at a large scale, and we are confident that it can contribute to bringing a better life in this pandemic.\\\\

\section{Future work}
For future work, COVID-19 social distancing with
person detection and tracking could be merged with this model to calculate the distance of people considering the factor whether personal protective equipment is used or not \cite{punn2020monitoring}.
\bibliographystyle{ieeetr}
\bibliography{sample.bib}







\end{document}